\documentclass[lettersize,journal]{IEEEtran}
\usepackage{amsmath,amsfonts}
\usepackage{algorithm}
\usepackage{array}
\usepackage[caption=false,font=normalsize,labelfont=sf,textfont=sf]{subfig}
\usepackage{textcomp}
\usepackage{stfloats}
\usepackage{url}
\usepackage{verbatim}
\usepackage{graphicx}
\usepackage{cite}
\usepackage{mdwmath}
\usepackage{eqparbox}
\usepackage{algpseudocode}
\usepackage{bm}
\usepackage{makecell}
\usepackage{pdflscape}
\usepackage{afterpage}
\usepackage{bbding}
\usepackage{lscape}
\usepackage{booktabs}
\usepackage{tikz}
\usepackage{bbding}
\usepackage{pgfplots}
\usepackage{rotating}
\usepackage{tabularray}
\usepackage{tabularx}
\usepackage{ragged2e}
\usepackage{threeparttable}
\usepackage[table]{xcolor}
\usepackage{multirow}
\usepackage[pagebackref=false,breaklinks=false,linkcolor=red,anchorcolor=black, citecolor=green,colorlinks,bookmarks=true]{hyperref}

\begin{document}

\title{OptiSAR-Net++: A Large-Scale Benchmark and Transformer-Free Framework for Cross-Domain Remote Sensing Visual Grounding}

\author{Xiaoyu Tang,~\IEEEmembership{Member,~IEEE,} Jun Dong,~\IEEEmembership{Student Member,~IEEE,} Jintao Cheng, Rui Fan,~\IEEEmembership{Senior Member,~IEEE}

\thanks{This work was partially supported by the National Natural Science Foundation of China under Grants 62473288 and 62233013, Guangdong Basic and Applied Basic Research Foundation (2024A1515012126), the Science and Technology Commission of Shanghai Municipal under Grant 22511104500, the Fundamental Research Funds for the Central Universities, and Xiaomi Young Talents Program. (Xiaoyu Tang and Rui Fan are co-corresponding authors).

Xiaoyu Tang is with the School of Electronics and Information Engineering, and Xingzhi College, South China Normal University, Foshan 528225, China. (e-mail: {\tt\small tangxy@scnu.edu.cn})

Jun Dong is with the School of Data Science and Engineering, and Xingzhi College, South China Normal University, Shanwei, 516600, China (e-mail: {\tt\small 20228132044@m.scnu.edu.cn}).

Jintao Cheng is with the Department of Electronic and Computer Engineering, Hong Kong University of Science and Technology, Hong Kong SAR, China. (e-mail: {\tt\small jchengau@connect.ust.hk})

Rui Fan is with the College of Electronic and Information Engineering, Shanghai Institute of Intelligent Science and Technology, Shanghai Research Institute for Intelligent Autonomous Systems, Shanghai Key Laboratory of Intelligent Autonomous Systems, State Key Laboratory of Autonomous Intelligent Unmanned Systems, and Frontiers Science Center for Intelligent Autonomous Systems (Ministry of Education), Tongji University, Shanghai 201804, China. (e-mail: {\tt\small rui.fan@ieee.org})

}
}

\markboth{Journal of \LaTeX\ Class Files,~Vol.~14, No.~8, August~2021}%
{Shell \MakeLowercase{\textit{et al.}}: A Sample Article Using IEEEtran.cls for IEEE Journals}

\IEEEpubid{0000--0000/00\$00.00~\copyright~2021 IEEE}

\maketitle

\begin{abstract}
Remote sensing visual grounding (RSVG) aims to localize specific targets in remote sensing images using natural language expressions. However, existing methods are restricted to single-sensor domains, i.e., either optical or synthetic aperture radar (SAR), limiting their real-world applicability. In this paper, we introduce the Cross-Domain RSVG (CD-RSVG) task and construct OptSAR-RSVG, the first large-scale benchmark dataset for this setting. To tackle the challenges of cross-domain feature modeling, computational inefficiency, and fine-grained semantic discrimination, we propose OptiSAR-Net++. Our framework features a patch-level Low-Rank Adaptation Mixture of Experts (PL-MoE) for efficient cross-domain feature decoupling. To mitigate the substantial computational overhead of Transformer decoding frameworks, we adopt a CLIP-based contrastive paradigm and further incorporate dynamic adversarial negative sampling, thereby transforming generative regression into an efficient cross-modal matching process. Additionally, a text-guided dual-gate fusion module (TGDF-SSA) and a region-aware auxiliary head are introduced to enhance semantic-visual alignment and spatial modeling. Extensive experiments demonstrate that OptiSAR-Net++ achieves SOTA performance on both OptSAR-RSVG and DIOR-RSVG benchmarks, offering significant advantages in localization accuracy and efficiency. 
The model and dataset have been made publicly available at
\url{https://github.com/JunDong-dev/OptiSAR-Net-PlusPlus}.
\end{abstract}

\begin{IEEEkeywords}
Visual grounding, foundation model, remote sensing, cross-domain image learning, contrastive learning.
% contrastive language-image pre-training.
\end{IEEEkeywords}

\section{Introduction}
\IEEEPARstart{T}{he} rapid advancement of remote sensing technology provides massive, high-precision earth observation data for urban planning~\cite{pami2}, environmental monitoring~\cite{pami1}, and national defense~\cite{saratr}. This creates a critical need for intelligent visual interpretation using natural language. Despite progress in tasks like image captioning and cross-modal retrieval~\cite{pami6_spectralgpt}, remote sensing visual grounding (RSVG) remains a challenging frontier. Unlike standard object detection, RSVG requires models to accurately localize specific targets within complex scenes by comprehending open-ended natural language queries~\cite {review}. This capability is vital for time-sensitive human-machine interactions, such as emergency response, demanding both precision and generalizability.
Existing RSVG research primarily focuses on single-source scenarios, treating sensor modalities as isolated domains. Mainstream datasets like DIOR-RSVG~\cite{dior_rsvg} and OPT-RSVG~\cite{opt_rsvg} rely solely on the rich textures of optical images, while efforts like TACMT~\cite{tacmt} explore SAR-based grounding. This fragmented paradigm overlooks the shared knowledge and complementary information across sensor domains, fundamentally limiting the robustness of systems deployed in diverse environments.

\begin{figure}[!tb]
\centering
\includegraphics[width=3.5in]{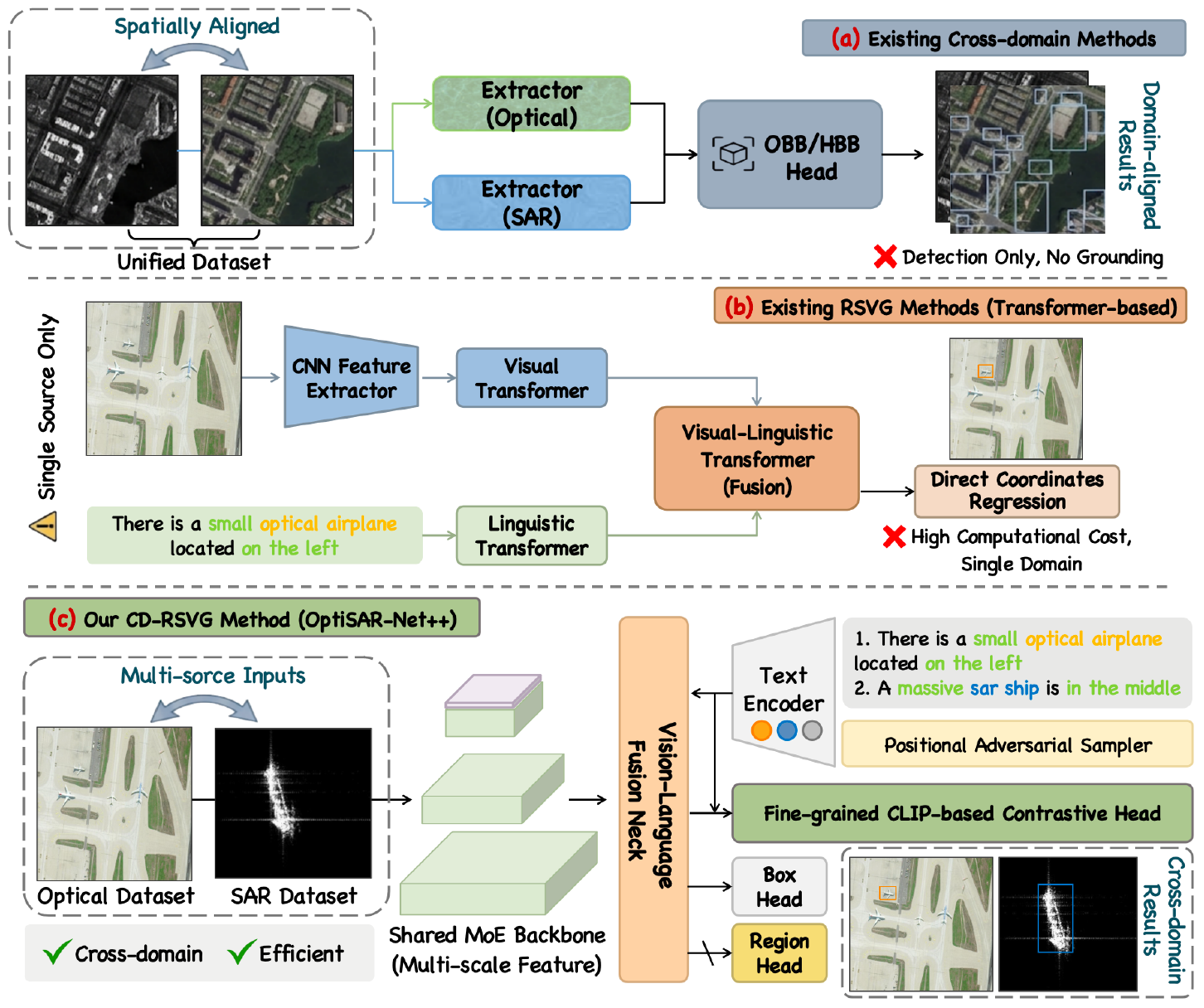}
\caption{Comparison between existing methods and the proposed CD-RSVG paradigm. (a) Existing cross-domain methods rely on co-registered image pairs and are mainly designed for detection, lacking language grounding. (b) Mainstream RSVG methods depend on heavy Transformers and single-source data, resulting in limited cross-domain generalization. (c) Our OptiSAR-Net++ employs a MoE backbone to uniformly handle multi-source inputs, integrates visual-linguistic features, and adopts a CLIP-style matching paradigm for efficient referring visual grounding.}
\label{fig:1}
\end{figure}

Practically, optical sensors offer rich semantics but are sensitive to weather and illumination, whereas SAR sensors provide all-weather, all-day imaging. While their complementary advantages are validated in object detection~\cite{9,optisarnet,11} (Fig.~\ref{fig:1}(a)), these methods rely on strictly co-registered image pairs, precluding direct extension to language grounding. To bridge this gap, we propose the Cross-Domain RSVG (CD-RSVG) task. It aims to establish a unified framework that leverages multi-source optical and SAR data to acquire universal cross-domain knowledge, enabling generalizable query grounding within a single training batch. However, CD-RSVG introduces formidable challenges beyond traditional single-source RSVG:

\begin{itemize}
    \item \textbf{Complexity of cross-domain feature modeling.} Optical and SAR images have fundamentally different imaging mechanisms: optical images feature clear edges and textures, while SAR images suffer from speckle noise and geometric distortions. Consequently, fully shared dense models face representation bottlenecks, struggling to accommodate both domains without negative transfer~\cite{sm3det}. This necessitates a parameter-efficient decoupling strategy to preserve domain-specific features while retaining cross-domain generalizability.  \newpage 

    \item \textbf{Computational efficiency bottleneck.} Existing RSVG methods predominantly rely on heavy Transformer architectures for multi-modal fusion (Fig.~\ref{fig:1}(b)), incurring substantial training costs and prohibitive inference latency. This poses critical deployment challenges for resource-constrained edge platforms, such as satellites and UAVs. There is an urgent need for efficient architectures that balance grounding accuracy with computational economy.
   
    \item \textbf{Inherent RSVG challenges in cross-domain settings.} Remote sensing targets exhibit extreme scale variations, posing fundamental challenges to fixed-receptive-field extractors. Furthermore, existing vision-language fusion mechanisms~\cite{14} typically embed directional information implicitly via cross-attention, lacking explicit alignment between directional semantics and visual features. This causes localization ambiguity in high-density scenes~\cite{csdnet}. Combined with the cross-domain setting, jointly modeling semantic commonalities and characteristic differences across modalities remains an unaddressed challenge.
\end{itemize}

To address these challenges, we construct \textbf{OptSAR-RSVG}, the first large-scale benchmark dataset for the CD-RSVG task, by consolidating and standardizing existing single-source datasets~\cite{vgrss,opt_rsvg,tacmt}. To mitigate class imbalance and limited semantic diversity, we apply geometric augmentation to under-represented categories and leverage GPT~\cite{gpt} to generate richer text descriptions. After rigorous manual verification, the final dataset comprises 46,825 images and 90,148 image-text-box triplet annotations. Specifically, the SAR subset contains 20,178 instances across 2 categories (12,466 images), while the optical subset contains 69,970 instances across 14 categories (34,359 images).

Furthermore, we extend the cross-domain object detection framework OptiSAR-Net~\cite{optisarnet} to propose \textbf{OptiSAR-Net++}, the first framework tailored for the CD-RSVG task (Fig.~\ref{fig:1}(c)). Specifically, we introduce a Patch-Level Low-Rank Adaptation Mixture of Experts (PL-MoE). This parameter-efficient decoupling strategy combines sparse routing with an adaptive cosine Top-K gating mechanism, achieving robust routing and load balancing to accommodate the divergent visual patterns of optical and SAR domains. Second, we adopt a CLIP-based~\cite{clip} contrastive learning paradigm for efficient visual grounding. Unlike standard open-vocabulary detection (OVD), we explicitly inject adversarial negative samples during training. This fine-grained, semantic-awareness mechanism compels the model to discriminate confusable directional and cross-domain information within a shared feature space. During inference, we leverage CLIP's cross-modal matching for discriminative region selection, transforming RSVG from a generative prediction task into a computationally efficient retrieval-based matching paradigm. Finally, we propose Text-Guided Dual-Gate Fusion with Spatial Shuffle Attention (TGDF-SSA) to efficiently inject linguistic semantics into the visual feature pyramid without computationally intensive attention operations. Additionally, a region-aware auxiliary head is introduced during training to supervise the target spatial distribution modeling explicitly. The main contributions of this work are summarized as follows:

\begin{itemize}
    \item We construct \textbf{OptSAR-RSVG}, the first large-scale benchmark for CD-RSVG. By expanding existing datasets via geometric augmentation and GPT-based rewriting, it effectively mitigates class imbalance and semantic scarcity. The dataset comprises 46,825 optical/SAR images and 90,148 image-text-box annotations across 16 categories.

    \item We propose \textbf{OptiSAR-Net++}, a lightweight, Transformer-free CD-RSVG framework. Its core components include: (i) PL-MoE for efficient cross-domain feature decoupling; (ii) a CLIP-based contrastive paradigm with adversarial sampling for fast cross-modal matching; and (iii) TGDF-SSA coupled with a region-aware auxiliary head to enhance linguistic-visual fusion and spatial learning.

    \item Extensive experiments demonstrate that OptiSAR-Net++ achieves SOTA performance on OptSAR-RSVG, outperforming previous models by 7.61\% and 0.32\% in optical and SAR inference, respectively. Furthermore, it exhibits superior data efficiency, directional awareness, and highly competitive generalization on the DIOR-RSVG.
\end{itemize}

The remainder of this paper is organized as follows. Sect.~\ref{sec:related_work} reviews related work. Sect.~\ref{sec:methods} details the proposed OptiSAR-Net++ framework, and Sect.~\ref{sec:dataset_construction} introduces the OptSAR-RSVG dataset. Sect.~\ref{sec:exp} validates the effectiveness of our method through comprehensive experiments. Finally, Sect.~\ref{sec:conclusion} concludes the paper with limitations and future directions.

% \vspace{-2pt}
\section{Related Work}
\label{sec:related_work}

\subsection{Remote Sensing Visual Grounding}
RSVG requires fine-grained cross-modal understanding to handle the scale heterogeneity, semantic complexity, and background clutter inherent in remote sensing scenarios. Existing frameworks are predominantly Transformer-based or LLM-based~\cite{review}.

Transformer-based methods dominate current RSVG research. Early works established benchmarks and basic fusion mechanisms~\cite{dior_rsvg,transvg}. Subsequent research addressed specific remote sensing challenges by incorporating spatial directional enhancement~\cite{dsevg}, mitigating attention drift~\cite{lpva}, enabling efficient sparse decoding~\cite{csdnet}, and adapting to SAR imagery~\cite{tacmt}. Recent advancements also include diffusion models for iterative refinement~\cite{lpqr_dvg} and the application of LLMs for enhanced semantic reasoning~\cite{skyeyegpt,earthgpt}. However, the quadratic complexity of Transformer architectures incurs substantial computational costs, severely hindering deployment on resource-constrained edge platforms.

Meanwhile, CLIP-based contrastive learning has excelled in open-vocabulary detection (OVD)~\cite{yoloe,yolo_world}, but struggles with the complex spatial relationships required in VG. While RemoteCLIP~\cite{remoteclip} adapts CLIP for remote sensing, it focuses on image-level tasks rather than region-level localization. General VG models like GLIP~\cite{glip} and Grounding DINO~\cite{grounding_dino} perform poorly on the extreme scales and bird's-eye-view geometry of remote sensing. Recently, Efficient Grounding DINO~\cite{efficient_dino} introduced open-set contrastive learning to RSVG; however, its massive parameter count (169.3M) remains prohibitive for edge devices, and it ignores cross-domain challenges. To address these gaps, this paper pioneers a fine-grained, semantic-aware contrastive learning mechanism for CD-RSVG. By utilizing adversarial negative sampling, our approach precisely discriminates confusing directional and cross-domain information within a shared feature space.

\subsection{Cross-Domain Image Learning}
Cross-domain learning integrates multi-source data to overcome the perceptual limitations of single modalities. In remote sensing, the collaborative analysis of optical and SAR imagery has proven effective for semantic segmentation~\cite{difference}, object detection~\cite{pami3} and change detection~\cite{macon}. Existing research primarily evolves along two lines: cross-domain feature fusion and feature adaptation.

In cross-domain fusion, methods focus on extracting complementary semantics from strictly registered or weakly aligned heterogeneous data~\cite{m4_sar,pami7_fusion,dpal}. Conversely, cross-domain adaptation targets the representation gap between unregistered source and target domains. This is typically achieved through unsupervised domain adaptation~\cite{highdan,macon} or by leveraging frozen vision-language models (e.g., CLIP) as stable semantic bridges for feature alignment~\cite{vlsda}.

Notably, most existing methods rely on spatially aligned data or target specific domain shifts, with limited exploration of semantically related but spatially unaligned joint training. Recently, frameworks like OptiSAR-Net~\cite{optisarnet} and SM3Det~\cite{sm3det} have explored unaligned cross-domain knowledge sharing for object detection. However, cross-domain learning remains completely unexplored in RSVG. This paper pioneers the CD-RSVG task, opening a new research direction for the unified and efficient intelligent interpretation of multi-source remote sensing data.

\section{Methodology}
\label{sec:methods}

\begin{figure*}[]
\centering
\includegraphics[width=\textwidth]{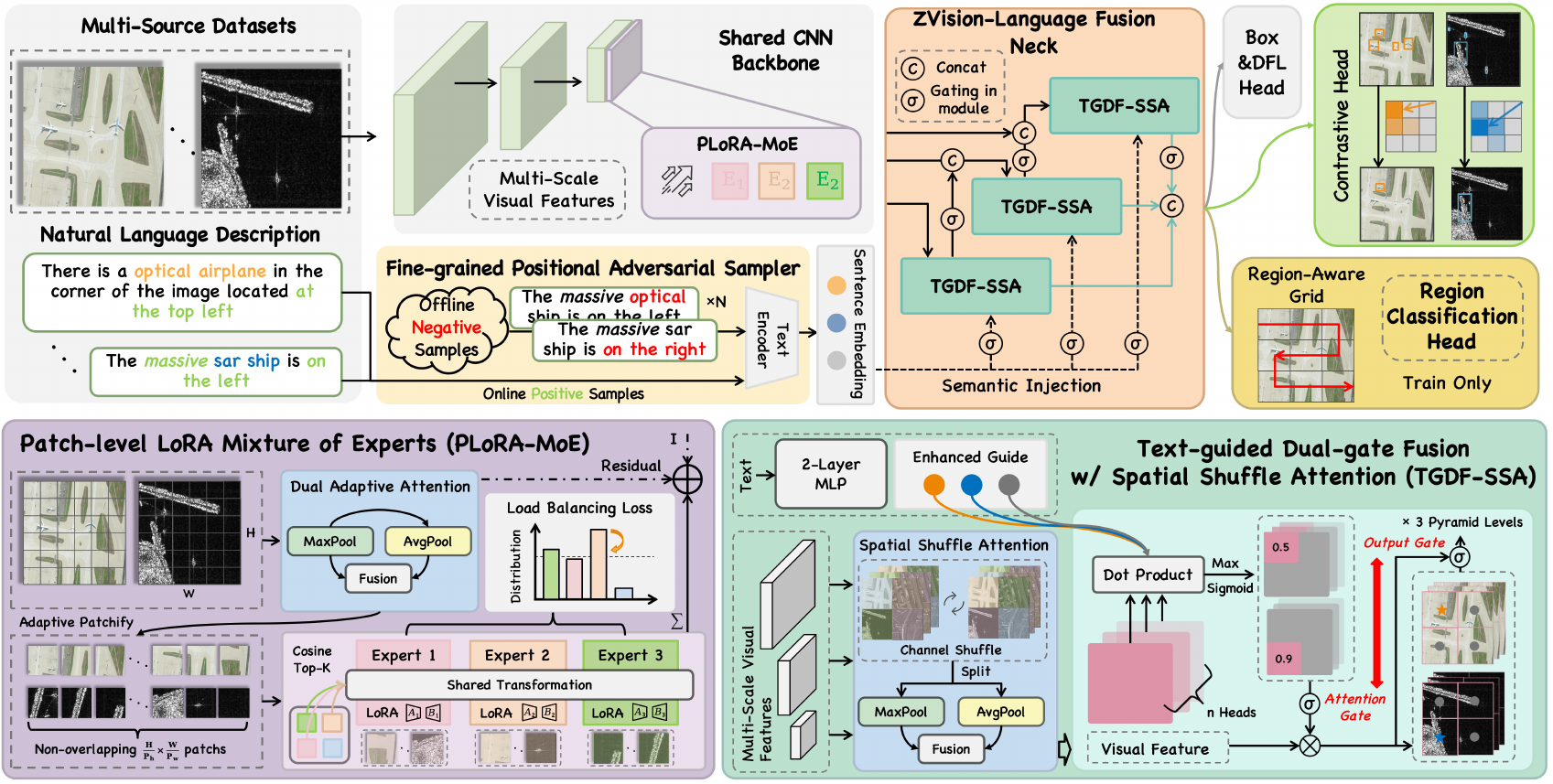}
\caption{Overall architecture of OptiSAR-Net++. The framework processes multi-source images and language queries for CD-RSVG via three main components: (1) A shared CNN backbone with PL-MoE, which adaptively routes image patches to domain-specific experts for cross-domain feature modeling. (2) A vision-language fusion neck (TGDF-SSA) that injects semantic information into multi-scale visual features. Text colors denote target categories (\textcolor{orange}{orange}/\textcolor{blue}{blue}) and attributes (\textcolor{green}{green}). (3) Detection heads comprising a regression head for candidate generation, a CLIP-based contrastive head for efficient retrieval matching, and an auxiliary region-aware classification head for spatial distribution modeling. During training, adversarial negatives (\textcolor{red}{red text}) are dynamically sampled to enhance fine-grained cross-domain grounding. Best viewed in color.}
\label{fig:overview}
\end{figure*}

\subsection{Preliminary}
In our prior work, OptiSAR-Net~\cite{optisarnet}, we introduced a cross-domain ship detection model for multi-source optical and SAR imagery. By incorporating attention mechanisms such as dual adaptive attention (DAA), bilevel routing deformable attention (BRDA), and spatial shuffling attention (SSA), it effectively focuses on target-related visual cues rather than variable backgrounds, enabling cross-domain generalization. 

However, OptiSAR-Net faces two main limitations. First, it is restricted to single-class ship detection and lacks natural-language understanding, limiting flexible human-computer interaction. Second, its dense, fully shared parameter architecture constrains its capacity for complex cross-domain feature modeling.
To address these issues, we propose \textbf{OptiSAR-Net++}, the first efficient Transformer-free framework tailored for CD-RSVG (Fig.~\ref{fig:overview}). Its advancements are threefold:
\textit{\textbf{(1) Task Upgrade.}} We extend the scope from predefined single-class detection to natural-language-guided visual grounding (VG) of general remote-sensing objects, enabling richer semantic understanding and precise localization.
\textit{\textbf{(2) Improved Cross-Domain Learning.}} We introduce a Patch-Level Low-Rank Adaptation Mixture of Experts (PL-MoE) module. Using a cosine-gating mechanism for sparse routing, it adaptively allocates expert networks to image patches from different domains, enabling efficient cross-domain feature modeling (Sect.~\ref{sec:plora}).
\textit{\textbf{(3) Architectural Optimization.}} Built upon CSPNet~\cite{cspnet} and PAN~\cite{pan}, we introduce a CLIP-based contrastive learning paradigm~\cite{clip} for VG. By leveraging adversarial negatives, a fine-grained, semantic-awareness mechanism explicitly disentangles object-localization cues from cross-domain semantics in the feature space. This converts heavy Transformer-based generative inference into efficient retrieval matching (Sect.~\ref{sec:clip}). Furthermore, our proposed TGDF-SSA module precisely injects text semantics into the visual feature pyramid. At the same time, a region-aware auxiliary head explicitly models target spatial distributions to achieve deep cross-modal fusion (Sect.~\ref{sec:fuse}).

\subsection{Cross-Domain Image Modeling}
\label{sec:plora}
The subtle common cues in geometric appearance and spatial structure (e.g., contours and local edge orientations) of cross-domain objects provide a fundamental basis for cross-domain modeling. However, due to the significant differences in imaging mechanisms between optical and SAR remote sensing, fully parameter-shared modeling aimed at extracting common knowledge often leads to feature interference in the latent space. On the one hand, shared representations are forced to accommodate multiple-source distributions, resulting in blurred decision boundaries. On the other hand, SAR-specific cues tend to conflict with optical texture semantics, thereby weakening the model's representational capacity and cross-domain generalization performance.

Inspired by the modern MoE paradigm~\cite{sm3det,modern_moe}, we propose PL-MoE to enhance cross-domain modeling capabilities while maintaining computational efficiency. As illustrated in Fig.~\ref{fig:overview}, PL-MoE incorporates two core architectural innovations: a patch-level sparse routing strategy and a parameter-efficient shared low-rank adaptation (LoRA)~\cite{lora} expert framework. By collaboratively modeling local visual patterns shared across modalities and allocating modality-specific cues to distinct experts for decoupled learning, our approach effectively alleviates negative transfer and mitigates representation conflicts.

\textit{\textbf{Patch-level Sparse Routing.}}
In contrast to routing the entire feature map to a single expert or employing pixel-level dense routing, we implement dynamic patch-level routing decisions on the feature map generated by the visual encoder to achieve an optimal balance between expressiveness and efficiency. Specifically, given an input feature map $\mathrm{X} \in \mathbb{R}^{B \times C \times H \times W}$, we partition it into $N = \frac{H}{p_h} \times \frac{W}{p_w}$ non-overlapping local patches ${\mathrm{P}_{ij}}$, where $p_h$ and $p_w$ denote adaptive patch dimensions. This patch-level routing mechanism enables experts to align with and process similar spatial patterns across domains, focusing on local structural patterns that are stable across modalities, thereby facilitating the extraction and reuse of shared knowledge.

\textit{\textbf{Shared LoRA Experts.}}
For each expert module, we adopt a hybrid architecture integrating a shared backbone with LoRA. All experts share a unified convolutional transformation $\mathcal{C}_{\mathrm{shared}}$ to preserve general representational capacity. Concurrently, each expert is augmented with a dedicated set of lightweight low-rank adapter matrices $\{(\mathrm{A}_e, \mathrm{B}_e)\}_{e=1}^{N_e}$, designed to capture expert-specific biases and facilitate fine-grained feature modulation, where $N_e$ denotes the total number of experts. The output of the $e$-th expert is formulated as follows:
\begin{equation}
\mathcal{E}_e(\mathrm{P})
=
\mathcal{C}_{\mathrm{shared}}(\mathrm{P})
+
\alpha \cdot \mathrm{P}\,\mathrm{A}_e\,\mathrm{B}_e,
\end{equation}
where $\mathrm{A}_e \in \mathbb{R}^{C \times r}$ and $\mathrm{B}_e \in \mathbb{R}^{r \times C}$ are the down- and up-projection matrices, respectively, $r \ll C$ is the low-rank dimension, and $\alpha$ is a learnable scaling factor.
To prevent expert collapse, we adopt a cosine similarity-based Top-$K$ gating mechanism and introduce a load balancing loss $\mathcal{L}_{\mathrm{aux}}$ to improve routing stability. Furthermore, we incorporate the DAA module from OptiSAR-Net~\cite{optisarnet} prior to MoE routing, which combines convolutional channel attention and spatial attention to enhance responses in critical regions, thereby improving cross-domain generalization. The complete output is formulated as:

\begin{equation}
\small
\mathrm{PLoRA\text{-}MoE}(\bar{\mathrm{P}})
=
\sum_{e=1}^{N_e}
g_e(\bar{\mathrm{P}})\,(\mathrm{P}\mathrm{A}_e\mathrm{B}_e).
\end{equation}

\begin{equation}
\small
\mathrm{Y}
=
\{(\mathrm{DAA}+\mathrm{PLoRA\text{-}MoE}+I)^{(k)}\}_{k=1}^{n}
\end{equation}
here, the visual features produced by the visual encoder are sequentially passed through $n$ cascaded modules, where the multi-scale contextual dependencies captured by $\mathrm{DAA}$ are fed into $\mathrm{PLoRA\text{-}MoE}$ for expert routing. After progressive feature refinement and the residual connection $\mathrm{I}$, the final output $\mathrm{Y}$ is generated. In $\mathrm{PLoRA\text{-}MoE}$, the sparse Top-$K$ gating function $g_e(\bar{\mathrm{P}})$ dynamically assigns routing weights based on the patch-level global feature $\bar{\mathrm{P}}$, enabling adaptive modeling of cross-domain features.

\subsection{CLIP-based Visual Grounding}
\label{sec:clip}

Most existing RSVG methods formulate localization as an autoregressive sequence generation problem~\cite{review}. However, their reliance on quadratic-complexity attention mechanisms incurs substantial computational overhead and low training efficiency. In contrast, CLIP-based contrastive learning~\cite{clip} has proven highly efficient for open-vocabulary detection (OVD)~\cite{grounding_dino,yolo_world}. Yet, its application to VG remains underexplored, primarily because standard CLIP-based OVD lacks mechanisms to explicitly capture fine-grained attributes and spatial orientation cues, hindering precise text--region alignment.

To address this, we cast RSVG as a discriminative retrieval problem within a dense prediction framework. Specifically, dense visual features are projected into a semantic space and matched against text embeddings to compute region--text scores. Concretely, given multi-scale pyramid features $\{F^l\}$, we first generate visual embeddings via a classification head:
\begin{equation}
E^l = g_{\text{vis}}(F^l) \in \mathbb{R}^{B \times D \times H_l \times W_l},
\end{equation}
where $D$ denotes the embedding dimension. On the text side, we use the CLIP text encoder to pre-encode and cache all complete candidate descriptions in the training set,
$T=\{t_k\}_{k=1}^{K}$ with $t_k \in \mathbb{R}^{D}$.
We then compute text--region similarity maps using a lightweight Batch-Normalization-based contrastive head, and calibrate the logits with a learnable scale $\gamma$ and bias term $b$. The resulting scores
$S^l \in \mathbb{R}^{B \times K \times H_l \times W_l}$
are directly used as the semantic classification logits. During inference, the multi-text contrastive formulation naturally degenerates into a single-query retrieval setting with one text embedding $t\in\mathbb{R}^{D}$, and we select the candidate region with the highest matching score as the final grounding prediction.

\textit{\textbf{Learning Strategy.}}
Unlike OVD training~\cite{yolo_world,yoloe} that typically assumes a fixed category set and applies forced contrastive sampling, RSVG queries depend critically on relational descriptions and spatial orientation. We therefore propose a fine-grained, semantic-aware adversarial negative sampling strategy. For each training instance, we generate adversarial textual variants according to predefined rules, in addition to constructing positive samples. Specifically, we create hard negatives by swapping orientation words (e.g., \textit{upper left} $\leftrightarrow$ \textit{upper right}) or cross-domain modality words (e.g., \textit{SAR} $\leftrightarrow$ \textit{optical}). Although these negatives are lexically close to the positives, they differ fundamentally in spatial reference or domain semantics, thereby forcing the model to explicitly reason about target orientation and modality cues and improving its discriminative capability for complex VG relations.

\textit{\textbf{Training Objective.}}
The overall objective consists of a regression loss $\mathcal{L}_{\text{reg}}$, a CLIP-based classification loss $\mathcal{L}_{\text{cls}}$, an auxiliary spatial loss $\mathcal{L}_{\text{region}}$, and the load-balancing regularizer $\mathcal{L}_{\text{lb}}$ introduced by PL-MoE. We adopt binary cross-entropy for $\mathcal{L}_{\text{cls}}$, while the regression term combines an IoU-based loss $\mathcal{L}_{\text{box}}$ with the distribution focal loss (DFL) $\mathcal{L}_{\text{dfl}}$.

\subsection{Explicit Visual-Linguistic Feature Fusion}
\label{sec:fuse}

While implicit visual-linguistic interaction via multi-layer cross-attention has become an effective cross-modal fusion approach in academia, this fusion process struggles to explicitly decouple the inherent spatially structured representations in remote sensing images. Such implicit coupling tends to introduce localization ambiguity in scenarios with high-density object distributions or strong background interference.

\textit{\textbf{TGDF-SSA.}} 
Inspired by~\cite{gate,yolo_world}, we propose TGDF-SSA, which precisely injects linguistic conditions into multi-scale local visual features in a controllable and interpretable manner. Specifically, visual features are first projected into a multi-head embedding space $\phi(V) \in \mathbb{R}^{B \times n_h \times d_h \times H \times W}$, where $d_h = e_c / n_h$. Meanwhile, text embeddings are linearly enhanced to obtain guidance vectors for each head $\psi(T) \in \mathbb{R}^{B \times N \times n_h \times d_h}$. Subsequently, visual-text similarity is computed for each spatial location, and the maximum value along the candidate text dimension is taken to obtain the optimal semantic matching response:

\begin{equation}
\small
A_{b,m,h,w} = \max_{n \in [1, N]} \langle \phi(V)_{b,m,:,h,w},\ \psi(T)_{b,n,m,:} \rangle.
\end{equation}

To enhance cross-modal learning stability, we introduce learnable temperature parameter $\tau$ and per-head bias $b_m$ to obtain calibrated attention logits:
\begin{equation}
\small
\tilde{A}_{b,m,h,w} = \frac{A_{b,m,h,w}}{\tau} + b_m, \qquad A^{\sigma} = \sigma(\tilde{A}) \cdot s,
\end{equation}
where $s$ is an optional learnable scaling factor and $\sigma(\cdot)$ is the Sigmoid activation function. Finally, adaptive learning through attention gating and feature gating enhances the model's cross-modal modeling capability, yielding semantically enhanced visual features $\tilde{V}$:
\begin{equation}
\small
\tilde{V} = V+\beta \cdot V \odot [\alpha \cdot (A^{\sigma} - \mathrm{1})].
\end{equation}

TGDF-SSA also inherits the design philosophy of SSA~\cite{optisarnet}. By perturbing and reorganizing features in advance along the channel dimension, it breaks the coupling in imaging-statistics-related features, encouraging the network to focus on more stable structural and shape cues and thereby improving cross-domain generalization in remote sensing.

\textit{\textbf{Region-Aware Auxiliary Head.}}
To further facilitate TGDF-SSA's learning of orientation and regional semantics, we introduce a region-aware auxiliary branch that is only activated during training. This branch classifies each location on the high-level feature map into predefined spatial grids, providing structured spatial supervision signals for the main branch. Given a feature layer $Z \in \mathbb{R}^{B \times C' \times H_l \times W_l}$, the prediction head outputs $P=h_{\text{grid}}(Z) \in \mathbb{R}^{B \times G \times H_l \times W_l}$. For feature map location $(i,j)$, its grid label is determined by the landing point of its corresponding original image center coordinates in the normalized plane:
\begin{equation}
\small
y_{i,j} = \mathrm{row}(i) \cdot c + \mathrm{col}(j), \quad y_{i,j} \in \{0, \dots, G-1\},
\end{equation}
where $\mathrm{row}(\cdot)$ and $\mathrm{col}(\cdot)$ are determined by the number of grid rows and columns and the feature point mapping relationship. The final spatial auxiliary loss is computed using pixel-wise cross-entropy:
\begin{equation}
\small
\mathcal{L}_{\text{region}} = \frac{1}{BHW} \sum_{b=1}^{B} \sum_{i=1}^{H} \sum_{j=1}^{W} \mathrm{CE}\big(P_{b,:,i,j},\ y_{i,j}\big).
\end{equation}

The region auxiliary head focuses on modeling spatial location priors (Where) during training, while TGDF-SSA concentrates on semantic content matching (What), ensuring reliable visual-linguistic fusion under complex relational descriptions and orientation constraints.

\section{Dataset Construction}
\label{sec:dataset_construction}

In this section, we introduce the OptSAR-RSVG dataset. Sect.~\ref{sec:dataset_pipeline} describes the dataset construction pipeline, while Sect.~\ref{sec:dataset_analysis} provides statistical analysis and visualization.

\subsection{OptSAR-RSVG: A Cross-Domain Remote Sensing Visual Grounding Benchmark}
\label{sec:dataset_pipeline}

\begin{figure*}[!tb]
\centering
\includegraphics[width=\textwidth]{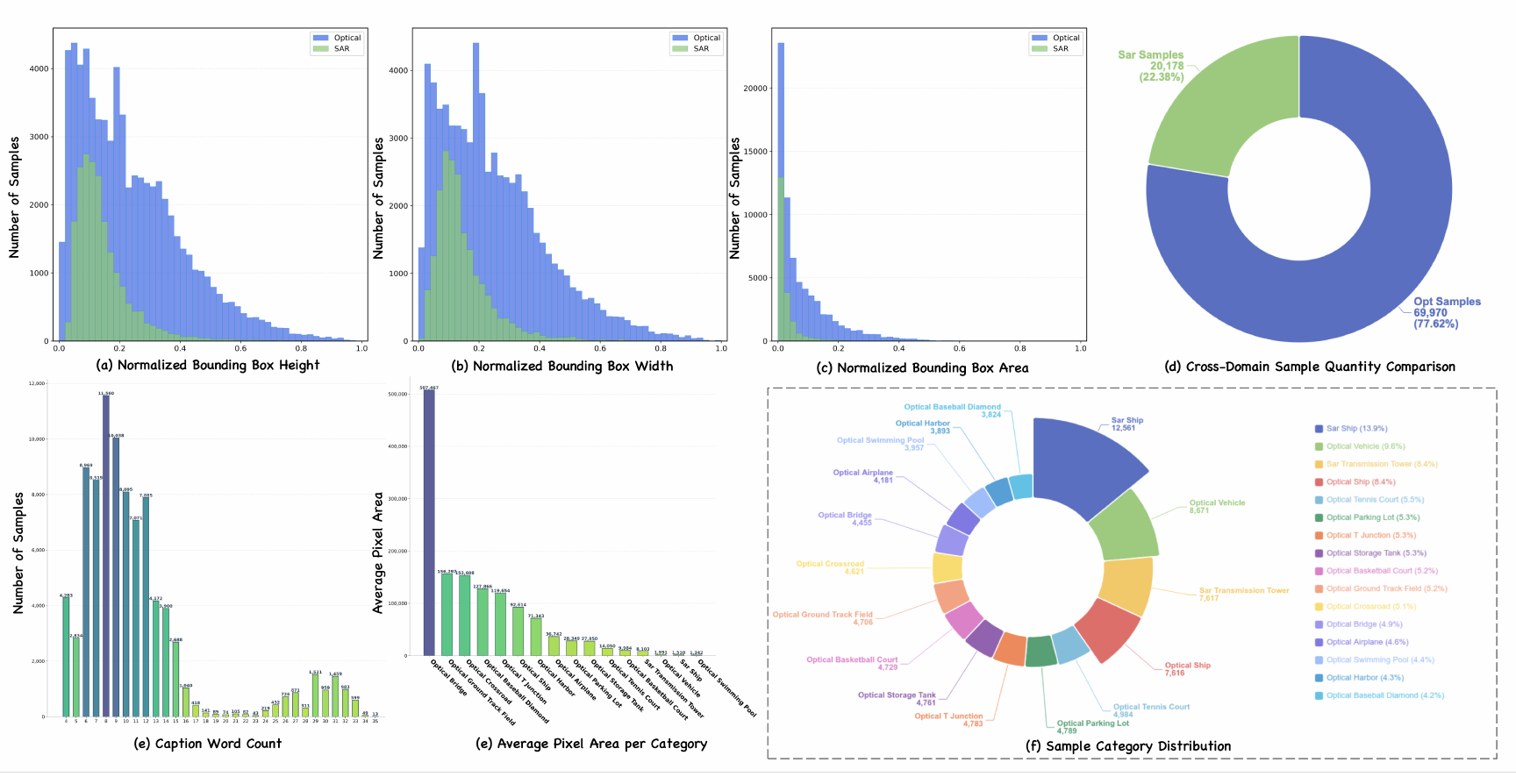}
\caption{Statistical analysis of the OptSAR-RSVG dataset. (a)-(c) Normalized bounding box height, width, and area distributions for optical (blue) and SAR (green) samples. Optical targets show broader scale variations, whereas SAR targets concentrate on medium-to-small scales. (d) Cross-domain sample quantity, reflecting practical data acquisition ratios. (e) Caption word count distribution (average: 11.13 words). (f) Average pixel area per category, highlighting inter-category scale diversity. (g) Sample distribution across 16 categories. Best viewed in color.}
\label{fig:data_analyze}
\end{figure*}

\begin{figure}[!tb]
\centering
\includegraphics[width=3.5in]{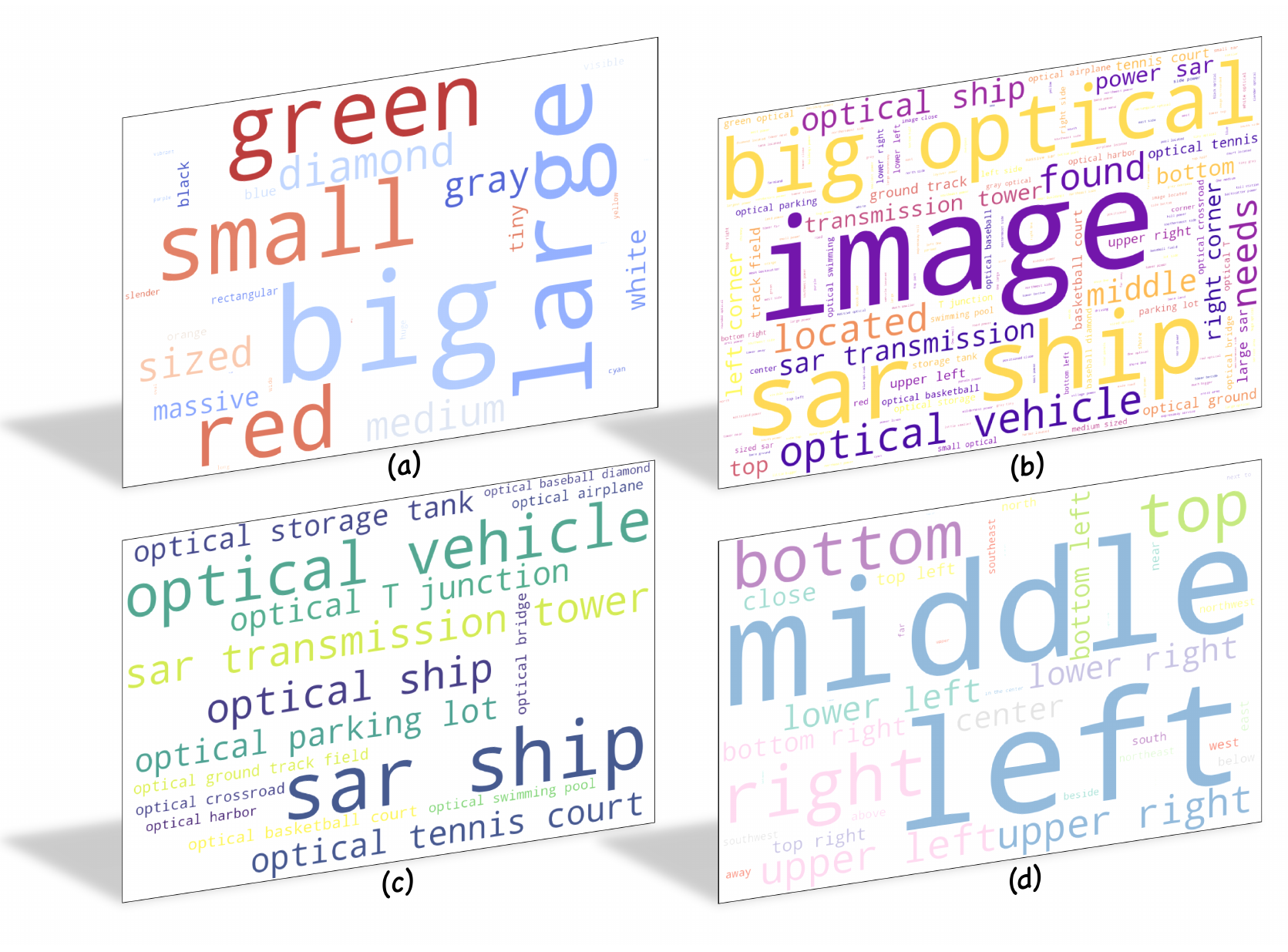}
\caption{Word cloud visualizations of textual descriptions in OptSAR-RSVG. (a) Target attributes, highlighting size and color descriptors. (b) Overall vocabulary, dominated by terms like "image", "ship", and modality identifiers. (c) Category names. (d) Spatial directional vocabulary, providing crucial semantic cues for localization.}
\label{fig:cloudworld}
\end{figure}

\begin{figure}[!tb]
\centering
\includegraphics[width=3.5in]{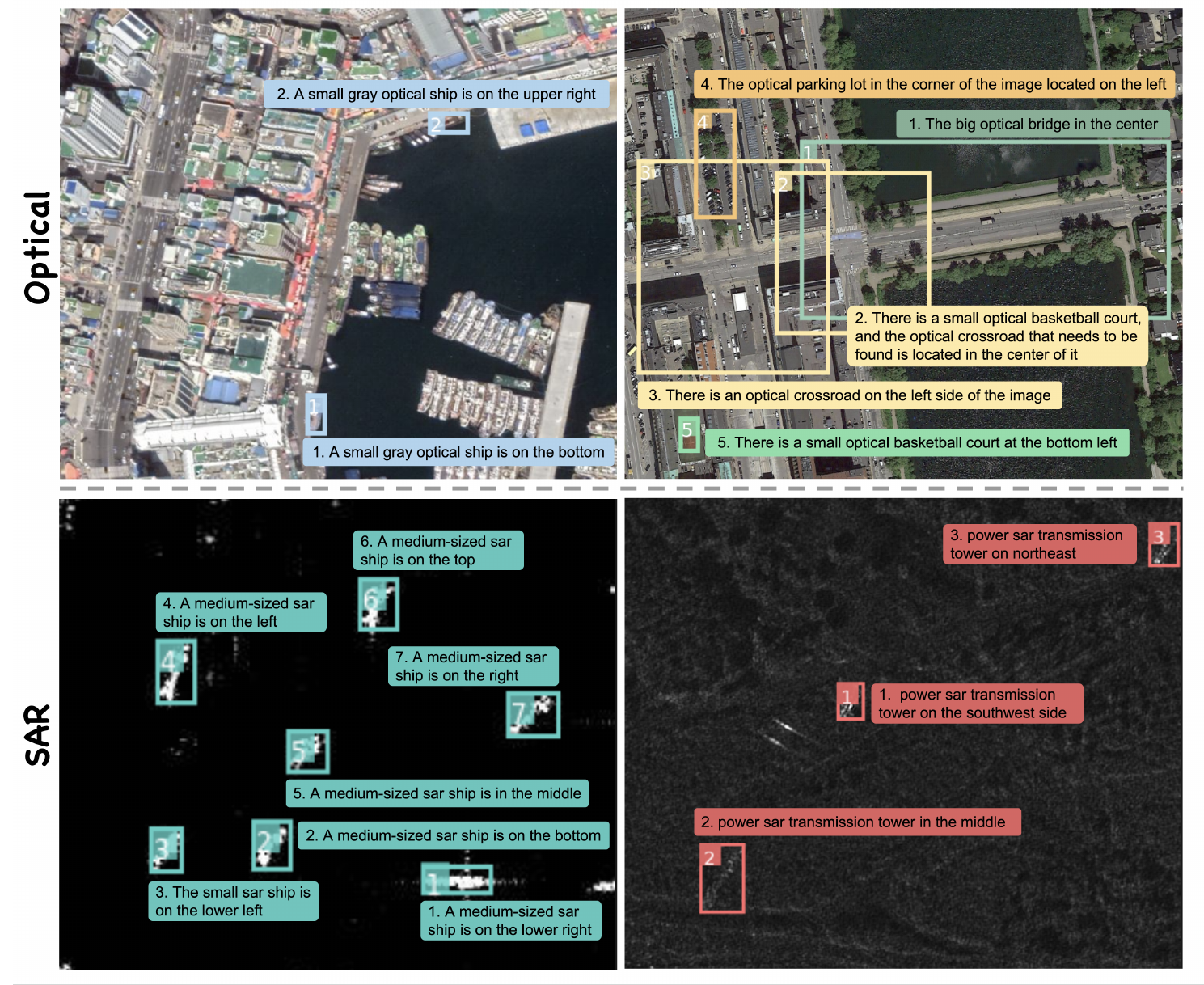}
\caption{Representative samples from the OptSAR-RSVG dataset, covering diverse optical (top row) and SAR (bottom row) scenarios. Each image is paired with a bounding box and a textual description containing target attributes, categories, and spatial cues. Optical samples exhibit rich semantic details, while SAR samples demonstrate target localization under challenging low-contrast conditions.}
\label{fig:optsar_rsvg}
\end{figure}

The CD-RSVG task requires remote sensing image data from both the optical and SAR domains, along with high-quality textual descriptions and bounding box annotations. However, existing RSVG datasets are constructed for single-source optical or SAR images, rendering them unsuitable for cross-domain research. To address this gap, we integrate four single-source RSVG datasets: RSVGD~\cite{vgrss}, OPT-RSVG~\cite{opt_rsvg} (which provides more abundant and complex annotations than DIOR-RSVG~\cite{dior_rsvg}), along with the only two publicly available SAR-based datasets, SARVG~\cite{vgrss} and TACMT~\cite{tacmt}. Through a systematic data processing pipeline, we construct OptSAR-RSVG, the first large-scale benchmark dataset for CD-RSVG tasks. The construction process comprises the following steps:

\textit{\textbf{Step 1 (Data Collection \& Cleaning):}} We collect images from four datasets: RSVGD~\cite{vgrss} (optical ships), OPT-RSVG~\cite{opt_rsvg} (14 optical targets), SARVG~\cite{vgrss} (SAR ships), and TACMT~\cite{tacmt} (SAR transmission towers). To ensure data quality, we first remove samples with invalid bounding boxes (e.g., out-of-bounds or \(x_{\min} \geq x_{\max}\) / \(y_{\min} \geq y_{\max}\)). Furthermore, we train an object detector on the merged data to identify annotation mismatches; images with a low predicted-to-ground-truth IoU are flagged for manual verification and potential removal. Finally, we unify the remaining 76 fine-grained ship categories in RSVGD into a single ``ship'' category (updating texts accordingly) and append modality prefixes (``Optical'' or ``SAR'') to all categories.

\textit{\textbf{Step 2 (Data Augmentation \& Merging):}} To mitigate class imbalance, we apply geometric augmentations (horizontal/vertical flips, 180-degree rotation) to minority classes, synchronously updating bounding boxes. We employ GPT-4o~\cite{gpt} to rewrite textual descriptions, updating absolute directional terms based on new coordinates while preserving relative positional semantics. An automated verification mechanism cross-checks the generated directional words against the transformed bounding box centers, prompting regeneration for conflicting samples. This process yields 9,046 new images with 19,136 annotations. The augmented data is then merged with the cleaned source datasets, resolving any filename conflicts via source-specific prefixes.

\textit{\textbf{Step 3 (Manual Verification \& Splitting):}} To eliminate potential semantic biases and ensure high-quality annotations, we conduct a final manual verification to check box-target matching and textual correctness. The finalized dataset is randomly split into training, validation, and test sets at an 8:1:1 ratio, maintaining consistent category and modality distributions. Specifically, the train/val/test sets contain 37,460/4,682/4,683 images and 72,118/9,015/9,015 annotations, respectively. Annotations are provided in both XML and COCO JSON formats.

\subsection{Statistical Analysis}
\label{sec:dataset_analysis}

We constructed the large-scale OptSAR-RSVG dataset, comprising 46,825 remote sensing images and 90,148 image/text description/bounding box triplet annotations spanning 16 target categories. The optical modality data includes 34,359 valid annotation files and 69,970 instances covering 14 categories; the SAR modality data includes 12,466 valid annotation files and 20,178 instances covering 2 categories.

\textit{\textbf{Cross-Domain Samples.}} Fig.~\ref{fig:data_analyze}(a)-(c) illustrate the width, height, and area distributions of the dataset. Optical targets exhibit a wide range of scales, whereas SAR targets (mainly ships and towers) are predominantly small to medium in size. As shown in Fig.~\ref{fig:data_analyze}(d), optical and SAR samples account for 77.6\% (69,970 instances) and 22.4\% (20,178 instances) of the dataset, respectively. This ratio reflects the real-world scarcity of SAR annotations while providing sufficient data to drive cross-domain learning research.

\textit{\textbf{Category Distribution.}} Fig.~\ref{fig:data_analyze}(g) presents the category proportions. Optical ships are abundant due to the merging of RSVGD~\cite{vgrss} and OPT-RSVG~\cite{opt_rsvg}, while other optical categories are balanced via augmentation. In the SAR domain, ships dominate over transmission towers. Furthermore, significant inter-class scale variations (Fig.~\ref{fig:data_analyze}(f)) demand robust multi-scale detection capabilities from visual grounding models. Category word clouds are visualized in Fig.~\ref{fig:cloudworld}(c).

\textit{\textbf{Textual Descriptions.}} Fig.~\ref{fig:data_analyze}(e) details the description length distribution. OptSAR-RSVG averages 11.13 words per description, surpassing existing datasets (e.g., DGF-RSVG~\cite{global} at 10.7, OPT-RSVG~\cite{opt_rsvg} at 10.10, RSVGD~\cite{vgrss} at 9.77, and DIOR-RSVG~\cite{dior_rsvg} at 7.47), indicating richer semantic expression. Word clouds for attributes, overall vocabulary, and directional words (Fig.~\ref{fig:cloudworld}(a, b, d)) highlight the dataset's linguistic diversity. Notably, the high frequency of directional terms emphasizes the importance of spatial cues for distinguishing similar remote sensing targets.

Fig.~\ref{fig:optsar_rsvg} presents sample data from the OptSAR-RSVG dataset, including typical images from both optical and SAR modalities. Each sample is annotated with target bounding boxes and corresponding textual descriptions. The samples demonstrate that the dataset covers diverse remote sensing scenarios, including urban areas, harbor waters, mountainous regions, and more.

\definecolor{aliceblue}{RGB}{240,248,255}

\section{Experiments}
\label{sec:exp}

In this section, we conduct extensive experiments to validate the effectiveness and generalizability of the proposed cross-domain remote sensing visual grounding framework. First, we introduce the experimental setup (Sect.~\ref{sec:setup}), covering datasets, evaluation metrics, and implementation details. Subsequently, we perform comprehensive comparative evaluations on our constructed OptSAR-RSVG dataset and the public single-source optical benchmark DIOR-RSVG (Sect.~\ref{sec:comparative}). Furthermore, we analyze the contributions of key components through ablation studies (Sect.~\ref{sec:ablation}). Finally, we present qualitative visualization results and demonstrate the unique zero-shot image annotation capability enabled by CLIP alignment (Sect.~\ref{sec:qualitative}).

\subsection{Experimental Setup}
\label{sec:setup}
\subsubsection{Datasets}
We evaluate the proposed method on two datasets. The first is OptSAR-RSVG, a large-scale CD-RSVG dataset we compiled that covers 16 common remote sensing object categories. Specifically, the optical subset comprises 34,359 valid annotation files with 69,970 instances across 14 categories, while the SAR subset comprises 12,466 annotation files with 20,178 instances across 2 categories. The dataset is randomly partitioned into training, validation, and test sets at 8:1:1, ensuring consistent category distributions and modal proportions across all subsets. To verify the generalizability of our model on standard benchmarks, we also conduct evaluations on DIOR-RSVG~\cite{dior_rsvg}, a published single-source optical dataset covering 20 object categories, with the training, validation, and test sets containing 26,991 (70\%), 3,829 (10\%), and 7,500 (20\%) image-text pairs, respectively.

\subsubsection{Evaluation Metrics}
Following the evaluation protocol of MGVLF~\cite{dior_rsvg}, we report Pr@0.5, Pr@0.6, Pr@0.7, Pr@0.8, Pr@0.9, meanIoU, and cumIoU. Let $\mathrm{IoU}(\hat{b}, b)$ denote the Intersection over Union between the predicted box $\hat{b}$ and the ground truth box $b$. The metric Pr@$t$ represents the percentage of test samples where the IoU is greater than or equal to the threshold $t$:
\begin{equation}
\small
\mathrm{Pr@}t = \frac{1}{N}\sum_{i=1}^{N}\mathbb{I}\big(\mathrm{IoU}(\hat{b}_i, b_i) \ge t\big) \times 100\%,
\end{equation}
where $\mathbb{I}(\cdot)$ is the indicator function. meanIoU is the average IoU over all test samples:
\begin{equation}
\small
\mathrm{meanIoU} = \frac{1}{N}\sum_{i=1}^{N}\mathrm{IoU}(\hat{b}_i, b_i) \times 100\%.
\end{equation}
cumIoU calculates the IoU by aggregating the intersection and union areas over the entire test set:
\begin{equation}
\small
\mathrm{cumIoU} = \frac{\sum_{i=1}^{N}\mathrm{Area}(\hat{b}_i \cap b_i)}{\sum_{i=1}^{N}\mathrm{Area}(\hat{b}_i \cup b_i)} \times 100\%.
\end{equation}

\begin{table*}[t]
    \caption{Performance comparison on OptSAR-RSVG benchmark. We evaluate our method against SOTA approaches across optical and SAR domains. Our OptSAR-Net++ achieves superior performance with significantly fewer parameters. The best two results are highlighted in \textcolor{red}{red} and \textcolor{blue}{blue}.}
    \label{tab:comparison-optsar}
    \renewcommand{\arraystretch}{1.0}
    \small
    \setlength{\tabcolsep}{2.5pt}
    \centering
    \resizebox{\linewidth}{!}{%
    \begin{tabular}{l|c|ccccc|ccccc|cc}
        \toprule
        \multirow{2}{*}{Method} &
          \multirow{2}{*}{\makecell{Params\\(M)}} &
          \multicolumn{5}{c|}{Optical Testset} &
          \multicolumn{5}{c|}{SAR Testset} &
          \multicolumn{2}{c}{All Testset} \\ 
        \cmidrule(lr){3-7} \cmidrule(lr){8-12} \cmidrule(lr){13-14}
         & & Pr@0.5 & Pr@0.7 & Pr@0.9 & meanIoU & cumIoU & Pr@0.5 & Pr@0.7 & Pr@0.9 & meanIoU & cumIoU & meanIoU & cumIoU \\ 
        \midrule
        \multicolumn{14}{l}{\textbf{\textit{Transformer-Based:}}} \\
        TransVG~\cite{transvg} & 149.7 & 43.06 & 30.10 & 3.23 & 36.87 & 39.86 & 65.99 & 41.47 & 1.64 & 51.72 & 20.04 & 40.69 & 40.90 \\
        LQVG~\cite{lqvg} & 156.8 & \textcolor{blue}{89.64} & \textcolor{blue}{80.06} & 33.48 & \textcolor{blue}{77.29} & 82.75 & \textcolor{blue}{94.13} & 87.39 & 32.14 & 80.76 & 81.04 & \textcolor{blue}{78.04} & 82.15 \\
        % LPVA~\cite{lpva} & 156.2 & 999 & 999 & 999 & 999 & 999 & 999 & 999 & 999 & 999 & 999 & 999 & 999 \\
        TACMT~\cite{tacmt} & 150.9 & 85.51 & 79.53 & \textcolor{blue}{42.20} & 75.74 & 79.98 & 92.87 & 89.74 & \textcolor{red}{38.29} & \textcolor{blue}{81.59} & \textcolor{blue}{82.46} & 77.24 & 81.40 \\
        CSDNet~\cite{csdnet} & 154.6 & 86.64 & 77.86 & 34.67 & 75.48 & \textcolor{blue}{83.20} & 93.55 & \textcolor{blue}{90.13} & 29.67 & 81.00 & 74.71 & 77.01 & \textcolor{blue}{82.96} \\
        \midrule
        \multicolumn{14}{l}{\textbf{\textit{Contrastive Learning-based:}}} \\
        G-DINO~\cite{grounding_dino} & 172.3 & 81.73 & 75.82 & 30.98 & 71.26 & 75.83 & 89.61 & 85.14 & 29.69 & 78.24 & 80.92 & 74.84 & 78.74 \\
        GLIP~\cite{glip} & 231.8 & 67.83 & 60.79 & 24.74 & 58.99 & 76.27 & 88.25 & 83.72 & 24.13 & 75.31 & 81.55 & 63.17 & 76.91 \\
        \midrule
        \rowcolor{cyan!10}
        OptSAR-Net++ (Ours) & 95.6 & \textcolor{red}{90.37} & \textcolor{red}{86.83} & \textcolor{red}{68.89} & \textcolor{red}{83.17} & \textcolor{red}{88.83} & \textcolor{red}{94.68} & \textcolor{red}{90.21} & \textcolor{blue}{37.22} & \textcolor{red}{81.85} & \textcolor{red}{83.15} & \textcolor{red}{82.76} & \textcolor{red}{90.70} \\
        \bottomrule
    \end{tabular}%
    }
    \vspace{2pt}
\end{table*}

\subsubsection{Implementation Details}
For the training of OptiSAR-Net++, we utilize the AdamW optimizer with an initial learning rate $\eta_{\text{init}} = 2 \times 10^{-3}$, weight decay of $0.025$, and momentum of $0.9$. A cosine learning rate decay strategy is adopted with a final factor of $0.01$. The warmup bias learning rate is set to $0.0$. The loss weights are configured as follows: region loss $\lambda_{\text{region}} = 1.0$, bounding box regression loss $\lambda_{\text{box}} = 7.5$, CLIP classification loss $\lambda_{\text{cls}} = 0.5$, DFL loss $\lambda_{\text{dfl}} = 1.5$, and the MoE load balancing loss weight is set to $\lambda_{\text{lb}} = 1.5$. The maximum number of text instances sampled per batch is set to 20. The sampling ratio of adversarial negative samples to positive samples is 0.5. The number of random negative samples is uniformly sampled from the interval $[0, 10]$, and global negative samples are used for padding to align tensor shapes.

For reproducing open-source methods, we strictly follow their official configurations, including loading pretrained visual backbones (e.g., ResNet-50, DETR) and text encoders (e.g., BERT). In contrast, our method only loads the pretrained MobileCLIP2-B~\cite{mobileclip2} text encoder for embedding generation. This branch contains 63.4M parameters and remains frozen throughout training without fine-tuning. \textbf{It is worth noting that, among the 95.6M total parameters of OptiSAR-Net++, only 32.2M parameters are trained from scratch and participate in training.} All experiments are conducted on 8 NVIDIA RTX 4090 GPUs using a distributed training strategy. OptiSAR-Net++ is trained for 300 epochs, while the other models are trained until the loss stops converging for 10 consecutive epochs. The global batch size is set to 64 for all models. The input image resolution is resized to $640\times640$ for both training and testing on OptSAR-RSVG and DIOR-RSVG~\cite{dior_rsvg}.

\subsection{Comparative Experiments}
\label{sec:comparative}

\begin{figure}[!tb]
\centering
\includegraphics[width=3.5in]{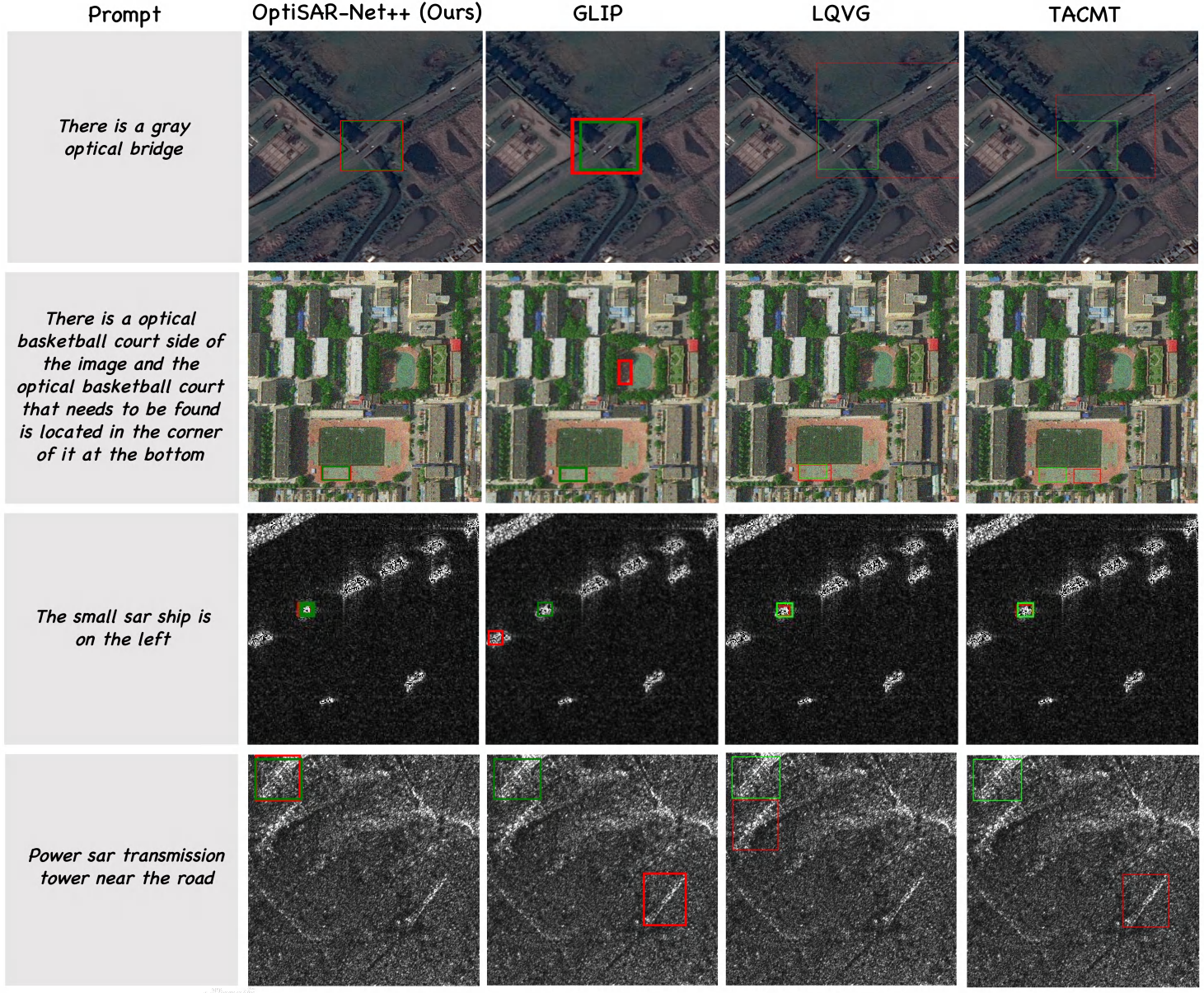}
\caption{CD-RSVG result comparison in optical and SAR scenes. \textcolor{green}{Green bounding boxes} denote the ground-truth annotations, and \textcolor{red}{red bounding boxes} indicate model predictions. The text prompts cover diverse semantic cues, including target category, attributes, and spatial location. OptiSAR-Net++ exhibits clear advantages in both localization accuracy and bounding-box regression quality.}
\label{fig:compare}
\end{figure}

This subsection validates the performance advantages of the proposed method through comprehensive comparisons. We first conduct a benchmark comparison with representative SOTA visual grounding methods on OptSAR-RSVG. Subsequently, we perform targeted evaluations from three perspectives: data efficiency, spatial localization capability, and generalizability on the public benchmark DIOR-RSVG.

\textit{\textbf{SOTA Comparisons on OptSAR-RSVG.}}
Table~\ref{tab:comparison-optsar} and Fig.~\ref{fig:compare} compare OptiSAR-Net++ with state-of-the-art methods. With only 95.6M parameters (including a frozen 63.4M MobileCLIP2 text branch~\cite{mobileclip2}), our model achieves the best overall test performance: 82.76\% meanIoU and 90.70\% cumIoU. It surpasses previous bests (LQVG: 78.04\% meanIoU; CSDNet: 82.96\% cumIoU) by 4.72\% and 7.74\%, respectively, while reducing the number of parameters by 36.1\%--58.8\% due to its efficient contrastive paradigm.

On the optical test set, OptiSAR-Net++ achieves 68.89\% Pr@0.9, outperforming TACMT (42.20\%) by 26.69\%, highlighting the precision of our mechanism's bounding-box regression. On the SAR test set, TACMT (38.29\%) slightly exceeds our method (37.22\%) on Pr@0.9. However, prior Transformer models struggle to balance cross-domain performance due to distribution shifts and imbalanced training data. Furthermore, despite larger capacities, contrastive baselines (Grounding DINO~\cite{grounding_dino}, GLIP~\cite{glip}) perform substantially worse, likely due to their reliance on natural-image pretraining without domain adaptation. Conversely, our domain-adaptive, hard-negative-enhanced CLIP paradigm effectively solves the CD-RSVG task.

\begin{figure}[!tb]
\centering
\includegraphics[width=3.5in]{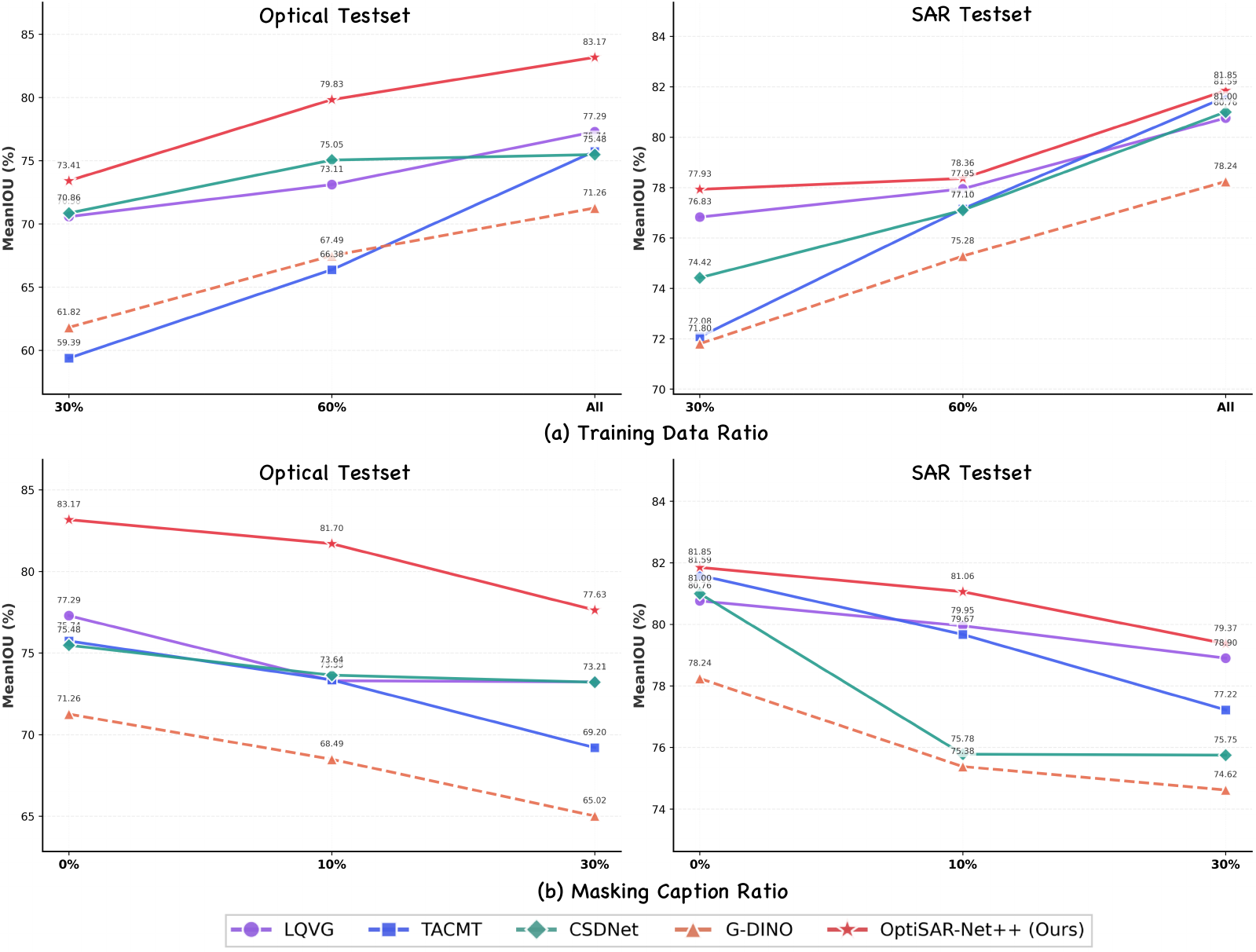}
\caption{Comparison of data efficiency on OptSAR-RSVG under different training data ratios (top) and comparison of spatial grounding performance under different description masking ratios (bottom).}
\label{fig:ratio}
\end{figure}

\textit{\textbf{Data Efficiency.}}
Table~\ref{tab:training-data} and Fig.~\ref{fig:ratio} evaluate meanIoU using 30\%, 60\%, and 100\% of the training data. OptiSAR-Net++ consistently outperforms competitors across all ratios. Notably, using only 60\% data, it achieves 79.83\% optical meanIoU, surpassing all full-data baselines (e.g., LQVG: 77.29\%). Remarkably, with just 30\% data, it attains 73.41\% (optical) and 77.93\% (SAR), rivaling or exceeding the full-data performance of much larger models like Grounding DINO (1.8$\times$ params; 71.26\%/78.24\%) and GLIP (2.4$\times$ params; 58.99\%/75.31\%).

This strong data efficiency stems from two factors: (1) the CLIP-style paradigm provides rich linguistic priors, reducing the data needed for cross-modal alignment; and (2) adversarial hard-negative sampling amplifies discriminative signals, forcing the model to extract fine-grained supervisory cues from limited samples.

\begin{table}[t]
\caption{Performance Comparison with Different Training Data (\%)}
\label{tab:training-data}
\centering
\scriptsize
\renewcommand{\arraystretch}{1.0}
\setlength{\tabcolsep}{3pt}
\begin{tabularx}{\linewidth}{@{}X|c|ccc|ccc@{}}
\toprule
\multirow{2}{*}{Method} & \multirow{2}{*}{\shortstack{Params\\(M)}} &
\multicolumn{3}{c|}{Optical Testset} & \multicolumn{3}{c}{SAR Testset} \\
\cmidrule(lr){3-5}\cmidrule(lr){6-8}
& & 30 & 60 & All & 30 & 60 & All \\
\midrule
\multicolumn{8}{l}{\textbf{\textit{Transformer-Based:}}} \\[2pt]
LQVG~\cite{lqvg}   & 156.8 & 70.58 & 73.11 & \textcolor{blue}{77.29} & \textcolor{blue}{76.83} & \textcolor{blue}{77.95} & 80.76 \\
TACMT~\cite{tacmt}  & 150.9 & 59.39 & 66.38 & 75.74 & 72.08 & 77.15 & \textcolor{blue}{81.59} \\
CSDNet~\cite{csdnet} & 154.6 & \textcolor{blue}{70.86} & \textcolor{blue}{75.05} & 75.48 & 74.42 & 77.10 & 81.00 \\
\midrule
\multicolumn{8}{l}{\textbf{\textit{Contrastive Learning-Based:}}} \\[2pt]
G-DINO~\cite{grounding_dino} & 172.3 & 61.82 & 67.49 & 71.26 & 71.80 & 75.28 & 78.24 \\
GLIP~\cite{glip}   & 231.8 & 49.88 & 56.02 & 58.99 & 68.71 & 73.44 & 75.31 \\
\midrule
\rowcolor{cyan!10}
OptSAR-Net++ (Ours) & 95.6 & \textcolor{red}{73.41} & \textcolor{red}{79.83} & \textcolor{red}{83.17} &
\textcolor{red}{77.93} & \textcolor{red}{78.36} & \textcolor{red}{81.85} \\
\bottomrule
\end{tabularx}
\end{table}

\textit{\textbf{Spatial Localization Capability.}}
A core challenge in RSVG is to accurately interpret orientation and spatial relations described in natural-language queries (e.g., ``vehicles on the north side of the playground''). To evaluate the spatial grounding capability of different methods, we randomly replace a portion of the textual descriptions with category names at ratios of 0\%, 10\%, and 30\%; the results are reported in Table~\ref{tab:masking-data}. On the optical test set, as the masking ratio increases from 0\% to 30\%, our meanIoU drops from 83.17\% to 77.63\%, a decrease of 5.54 percentage points (a relative drop of 6.66\%). On the SAR test set, the degradation is notably smaller, decreasing from 81.85\% to 79.37\%, i.e., only 2.48 percentage points (a relative drop of 3.03\%). In comparison, GLIP and TACMT exhibit the most severe performance degradation on the optical test set, indicating that their spatial reasoning is more fragile and that they are more susceptible to missing explicit orientation cues in the text queries. In contrast, the moderate degradation of OptiSAR-Net++ suggests that it leverages spatial and category cues in a balanced manner rather than over-relying on either.

\begin{table}[t]
\caption{Performance Comparison with Different Masking Caption Ratio (\%)}
\label{tab:masking-data}
\centering
\scriptsize
\setlength{\tabcolsep}{3pt}
\renewcommand{\arraystretch}{1.0}
\begin{tabularx}{\linewidth}{@{}X|c|ccc|ccc@{}}
\toprule
\multirow{2}{*}{Method} & \multirow{2}{*}{\shortstack{Params\\(M)}} &
\multicolumn{3}{c|}{Optical Testset} & \multicolumn{3}{c}{SAR Testset} \\
\cmidrule(lr){3-5}\cmidrule(lr){6-8}
& & 0 & 10 & 30 & 0 & 10 & 30 \\
\midrule
\multicolumn{8}{l}{\textbf{\textit{Transformer-Based:}}} \\[2pt]
LQVG~\cite{lqvg}   & 156.8 & \textcolor{blue}{77.29} & 73.30 & \textcolor{blue}{73.23} & 80.76 & \textcolor{blue}{79.95} & \textcolor{blue}{78.90} \\
TACMT~\cite{tacmt}  & 150.9 & 75.74 & 73.35 & 69.20 & \textcolor{blue}{81.59} & 79.67 & 77.22 \\
CSDNet~\cite{csdnet} & 154.6 & 75.48 & \textcolor{blue}{73.64} & 73.21 & 81.00 & 75.78 & 75.75 \\
\midrule
\multicolumn{8}{l}{\textbf{\textit{Contrastive Learning-Based:}}} \\[2pt]
G-DINO~\cite{grounding_dino} & 172.3 & 71.26 & 68.49 & 65.02 & 78.24 & 75.38 & 74.62 \\
GLIP~\cite{clip}   & 231.8 & 58.99 & 56.23 & 51.00 & 75.31 & 73.92 & 71.43 \\
\midrule
\rowcolor{cyan!10}
OptSAR-Net++ (Ours) & 95.6 & \textcolor{red}{83.17} & \textcolor{red}{81.70} & \textcolor{red}{77.63} &
\textcolor{red}{81.85} & \textcolor{red}{81.06} & \textcolor{red}{79.37} \\
\bottomrule
\end{tabularx}
\end{table}

\begin{table}[t]
\caption{Performance comparison on DIOR-RSVG benchmark.}
\label{tab:performance-comparison}
\centering
\footnotesize
\setlength{\tabcolsep}{4pt}
\renewcommand{\arraystretch}{1.0}

% --- column helpers for tabularx ---
\newcolumntype{Y}{>{\centering\arraybackslash}X} % centered X
\newcolumntype{L}{>{\raggedright\arraybackslash}l} % (optional) keep method left

\begin{tabularx}{\columnwidth}{L|c|YYYYY}
\toprule
Method &
\thead{Params\\(M)} &
Pr@0.5 &
Pr@0.7 &
Pr@0.9 &
\makecell{mean\\IoU} &
\makecell{cum\\IoU} \\
\midrule

\multicolumn{7}{l}{\textit{One-Stage Dense Prediction:}} \\
ZSGNet~\cite{zsgnet} & - & 51.67 & 42.30 & 10.15 & 44.12 & 51.65 \\
FAOA~\cite{faoa} & - & 70.86 & 62.04 & 36.44 & 62.86 & 67.28 \\
ReSC~\cite{resc} & 179.9 & 72.71 & 63.01 & 33.37 & 64.24 & 68.10 \\
LBYL-Net~\cite{lbyl} & 163.8 & 73.78 & 65.36 & 19.52 & 65.23 & 76.37 \\
\midrule

\multicolumn{7}{l}{\textit{Transformer-Based:}} \\
QRNet~\cite{qrnet} & 150.5 & 74.38 & 63.47 & 20.83 & 62.73 & 71.69 \\
TransVG~\cite{transvg} & 149.7 & 72.41 & 60.05 & 16.30 & 60.95 & 70.17 \\
MGVLF~\cite{dior_rsvg} & 152.5 & 76.78 & 66.74 & 35.07 & 68.04 & 78.41 \\
LQVG~\cite{lqvg} & 156.8 & 83.41 & \textcolor{red}{75.91} & 33.74 & \textcolor{red}{74.02} & 82.22 \\
LPVA~\cite{lpva} & 156.2 & 82.27 & 72.25 & 39.55 & 72.23 & \textcolor{blue}{85.11} \\
TACMT~\cite{tacmt} & 150.9 & 79.92 & 71.43 & 38.75 & 71.64 & 80.64 \\
MSANet~\cite{msa} & - & 74.23 & 61.32 & 24.26 & 64.88 & 77.13 \\
CSDNet~\cite{csdnet} & 154.6 & 81.91 & 71.05 & 32.04 & 70.88 & 79.58 \\
QAMFN~\cite{qamfn} & 128.4 & 81.67 & 68.56 & 34.51 & 71.78 & 84.55 \\
GLAED~\cite{glaed} & 157.2 & \textcolor{red}{84.44} & 73.51 & \textcolor{blue}{43.36} & 73.69 & 82.29 \\
\midrule

\multicolumn{7}{l}{\textit{Contrastive Learning:}} \\
Eff.\ G-DINO~\cite{efficient_dino} & 169.3 & 83.05 & 75.00 & 42.27 & 73.41 & 81.06 \\
G-DINO~\cite{grounding_dino} & 172.3 & 80.07 & 72.77 & 40.37 & 71.21 & 80.49 \\
GLIP~\cite{glip} & 231.8 & 76.14 & 70.93 & 35.87 & 69.86 & 79.05 \\
\midrule

\rowcolor{cyan!10}
OptSAR-Net++ (Ours) & 95.6 &
\textcolor{blue}{84.07} & \textcolor{blue}{75.69} &
\textcolor{red}{45.28} & \textcolor{blue}{73.79} &
\textcolor{red}{85.46} \\
\bottomrule
\end{tabularx}
\end{table}

\textit{\textbf{Generalization on DIOR-RSVG.}}
To assess the generalization of our framework beyond the cross-domain setting, we further evaluate it on the widely used single-source optical benchmark DIOR-RSVG~\cite{dior_rsvg}. As shown in Table~\ref{tab:performance-comparison}.

OptiSAR-Net++ achieves the best results on the two most challenging metrics: Pr@0.9 (45.28\%) and cumIoU (85.46\%), surpassing the runner-up GLAED~\cite{glaed} (43.36\%) by 1.92 percentage points and LPVA~\cite{lpva} (85.11\%) by 0.35 percentage points, respectively. On Pr@0.5 (84.07\%), Pr@0.7 (75.69\%), and meanIoU (73.79\%), our method ranks second, trailing GLAED (84.44\%), LQVG~\cite{lqvg} (75.91\%), and LQVG (74.02\%) by narrow margins of 0.37, 0.22, and 0.23 percentage points, respectively. These results indicate that, although OptiSAR-Net++ is designed for the cross-domain CD-RSVG task, it can effectively generalize to the single-source scenario without any architectural modifications, validating the generality of our framework.

\begin{figure}[!tb]
\centering
\includegraphics[width=3.5in]{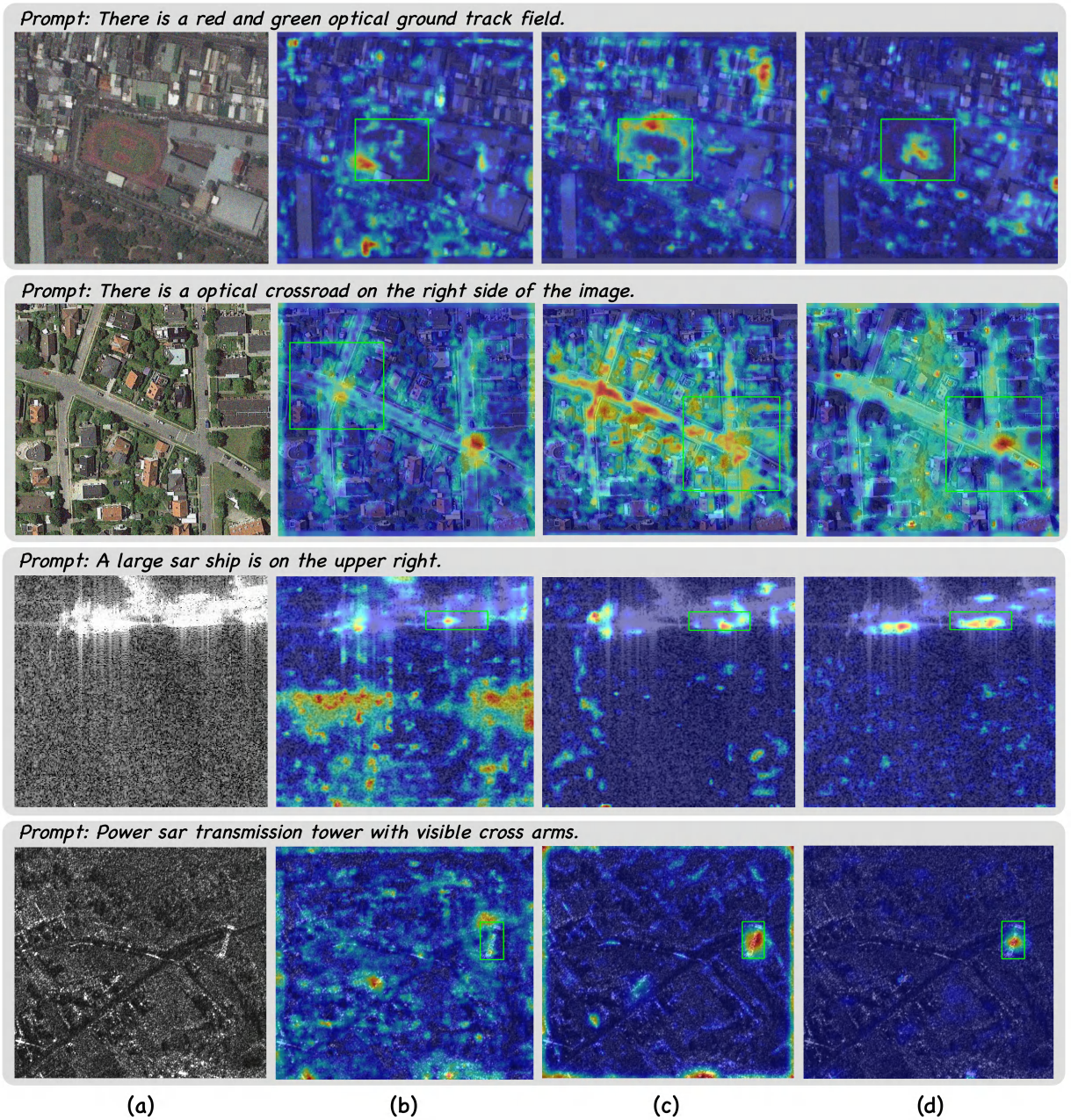}
\caption{Visualization of last-layer feature activation heatmaps across ablation stages for optical (top two rows) and SAR (bottom two rows) samples. (a) Input image. (b) Baseline model, showing dispersed activations across the scene. (c) Model with fine-grained adversarial sampling, exhibiting improved attention concentration. (d) Full OptiSAR-Net++ (integrating PL-MoE, TGDF-SSA, and the auxiliary head), yielding highly focused activations that precisely align with target boundaries. \textcolor{green}{Green boxes} denote localization results.}
\label{fig:heatmap}
\end{figure}

\begin{table}[t]
\caption{Ablation Study of Key Components}
\label{tab:ablation}
\centering
\small
\renewcommand{\arraystretch}{1.0}
\begin{tabularx}{\columnwidth}{ccc|c|XX}
\toprule
\makecell{Fine.\\Sample} & \makecell{PL-\\MoE} & \makecell{Sem.\\ Injection} & \makecell{Params\\(M)} & \makecell{mean\\IoU} & \makecell{cum\\IoU} \\
\midrule
 & & & 91.0 & 76.42 & 82.15 \\
\checkmark & & & 91.0 & 81.59$_{+5.17}$ & 88.03$_{+7.88}$ \\
\checkmark & \checkmark & & 91.3 & 82.34$_{+0.75}$ & 88.96$_{+0.93}$ \\
\rowcolor{cyan!10}
\checkmark & \checkmark & \checkmark & 95.6 & \textbf{82.76}$_{+0.42}$ & \textbf{90.70}$_{+1.74}$ \\
\bottomrule
\end{tabularx}
\end{table}

\subsection{Ablation Studies}
\label{sec:ablation}

In this subsection, we conduct ablation studies to analyze the contributions of key components to the overall performance, including ablations on core modules and MoE configurations.

\textit{\textbf{Ablation on Key Components.}}
Table~\ref{tab:ablation} details the progressive integration of each component into our baseline (a standard backbone+neck detector using CLIP-style random sampling; activations in Fig.~\ref{fig:heatmap}(b)). 
Adversarial negative sampling (perturbing orientation and domain cues) enables fine-grained feature distinction, shifting activations from dispersed to semantically focused (Fig.~\ref{fig:heatmap}(c)). Adding PL-MoE further improves meanIoU by 0.75\% and cumIoU by 0.93\% with only 0.3M extra parameters. This sparse architecture efficiently strengthens cross-domain representations by adaptively routing patches to domain-matched experts. Finally, the semantic injection module (TGDF-SSA and auxiliary head) yields further gains of 0.42\% meanIoU and 1.74\% cumIoU. The full model produces highly concentrated activations (Fig.~\ref{fig:heatmap}(d)), demonstrating the strong synergy among all proposed components.

\begin{figure}[!tb]
\centering
\includegraphics[width=3.5in]{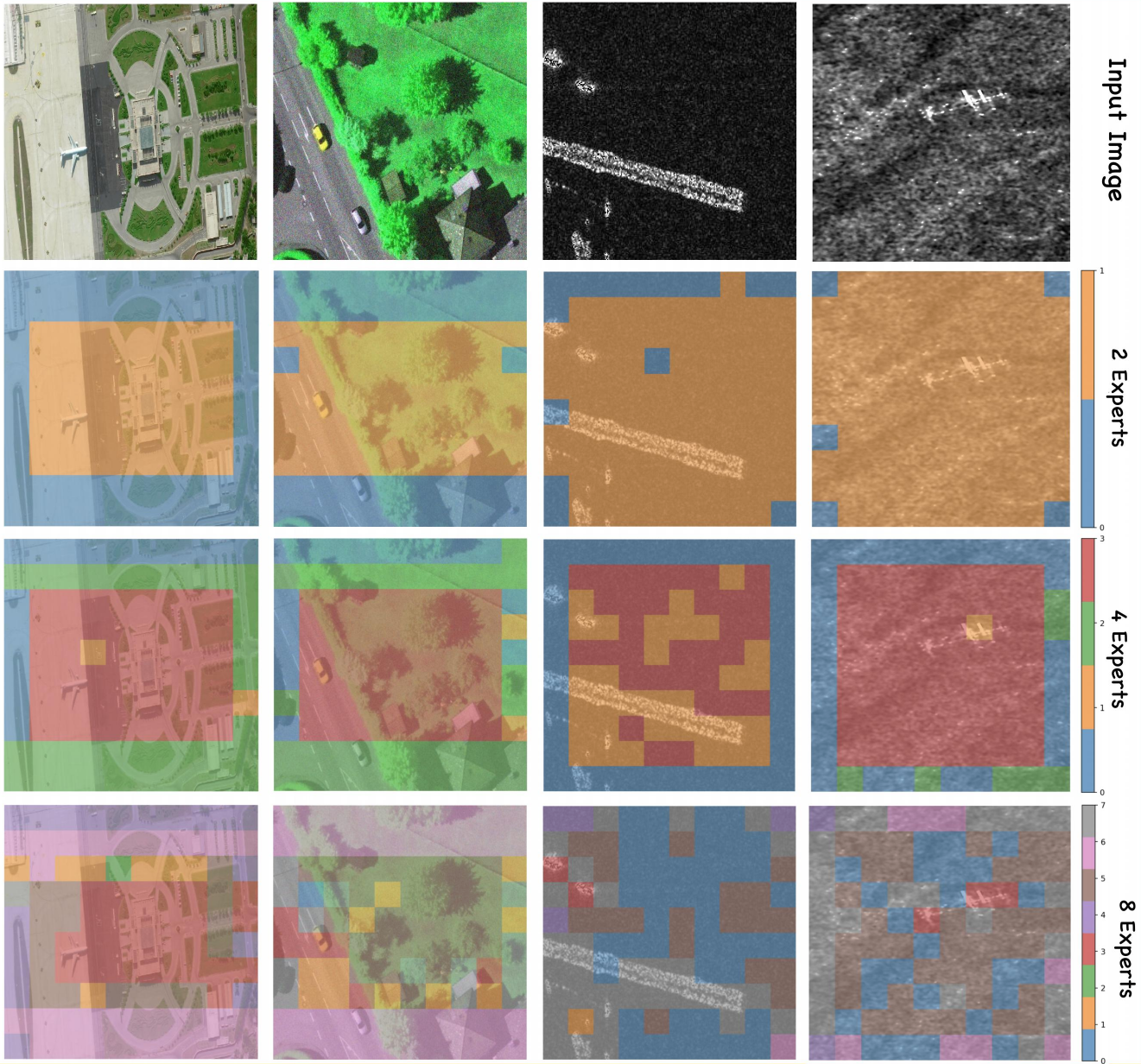}
\caption{Visualization of expert routing patterns under different MoE granularity settings. From top to bottom, the models use 2, 4, and 8 experts for cross-domain learning, respectively.}
\label{fig:moe}
\end{figure}

\textit{\textbf{Ablation on MoE Configurations.}}
Table~\ref{tab:moe-ablation} and Fig.~\ref{fig:moe} explore MoE configurations: expert count (\(n\)), Top-\(k\) selection (\(k\)), and routing granularity defined by patch size (\(p\)). 
With \(k\!=\!2\) and \(p\!=\!2\), increasing \(n\) from 2 to 8 consistently improves meanIoU (81.04\% \(\rightarrow\) 81.48\% \(\rightarrow\) 82.76\%), as expert diversity facilitates finer optical/SAR specialization. We set \(n\!=\!8\) to balance performance and efficiency. For Top-\(k\) selection, \(k\!=\!2\) is optimal: \(k\!=\!1\) limits the integration of shared representations, while \(k\!>\!2\) dilutes expert specialization. Regarding routing granularity, image-level routing (\(p\!=\!\infty\)) performs worst (81.37\% meanIoU, 87.96\% cumIoU). Token-level routing (\(p\!=\!1\)) achieves competitive meanIoU (82.73\%) but lower cumIoU (89.84\%) than patch-level routing (\(p\!=\!2\): 82.76\%/90.70\%). This suggests that overly fine-grained routing disrupts spatial coherence. Thus, \(p\!=\!2\) strikes the best balance, grouping neighboring tokens to preserve both spatial consistency and routing flexibility.

\begin{table}[t]
\caption{Ablation Study of MoE Configuration}
\label{tab:moe-ablation}
\centering
\footnotesize
\setlength{\tabcolsep}{2pt}
\resizebox{\columnwidth}{!}{%
\begin{tabular}{l|cc|cc|cccc}
\toprule
\makecell{MoE\\(n, k, p)} & \makecell{2, 2, 2} & \makecell{4, 2, 2} & \makecell{8, 1, 2} & \makecell{8, 3, 2}  & \makecell{8, 2, 1\\(G. Lv.)}  & \makecell{8, 2, $\infty$\\(I. Lv.)} & \makecell{8, 2, 4\\(P. Lv.)} & \makecell{8, 2, 2\\(P. Lv.)} \\
\midrule
Params (M) & 94.9 & 95.6 & 95.6 & 95.6 & 95.6 & 95.6 & 95.6 & 95.6\\
\cmidrule{1-9}
meanIoU & 81.04 & 81.48 & 81.97 & 82.64 & 82.73 & 81.37 & 82.71 & \textbf{82.76}\\
cumIoU & 87.76 & 89.82 & 88.56 & 89.35 & 89.84 & 87.96 & 90.48 & \textbf{90.70}\\
\bottomrule
\end{tabular}%
}
\end{table}

\begin{figure}[!tb]
\centering
\includegraphics[width=3.5in]{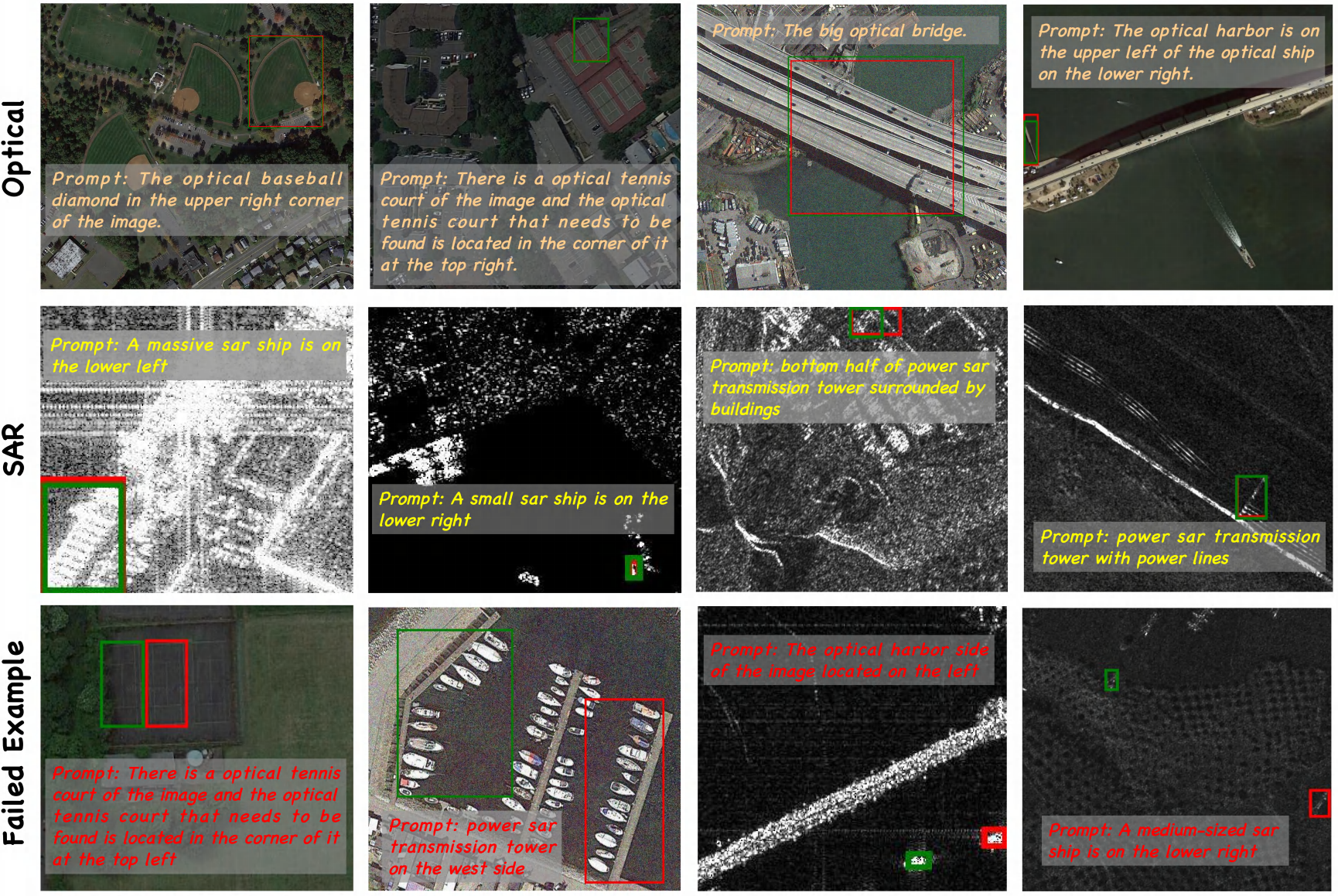}
\caption{Additional visual grounding predictions of OptiSAR-Net++ on the OptSAR-RSVG test set, covering a variety of representative scenes in optical (first row) and SAR (second row) imagery, as well as several failure cases (third row). For each example, the text query is shown next to the image; \textcolor{green}{green bounding boxes} denote the ground-truth annotations, and \textcolor{red}{red bounding boxes} indicate the predicted results.}
\label{fig:moredemo}
\end{figure}

\subsection{Further Qualitative Analysis}
\label{sec:qualitative}
We visualize the localization predictions of our method in more complex scenarios, such as dense object clusters, strong background interference, and cases with ambiguous descriptions. Additionally, we demonstrate the zero-shot image annotation capability enabled by the CLIP-aligned latent space, highlighting the semantic generalization potential of our framework for future data annotation tasks.

\begin{figure}[]
\centering
\includegraphics[width=3.5in]{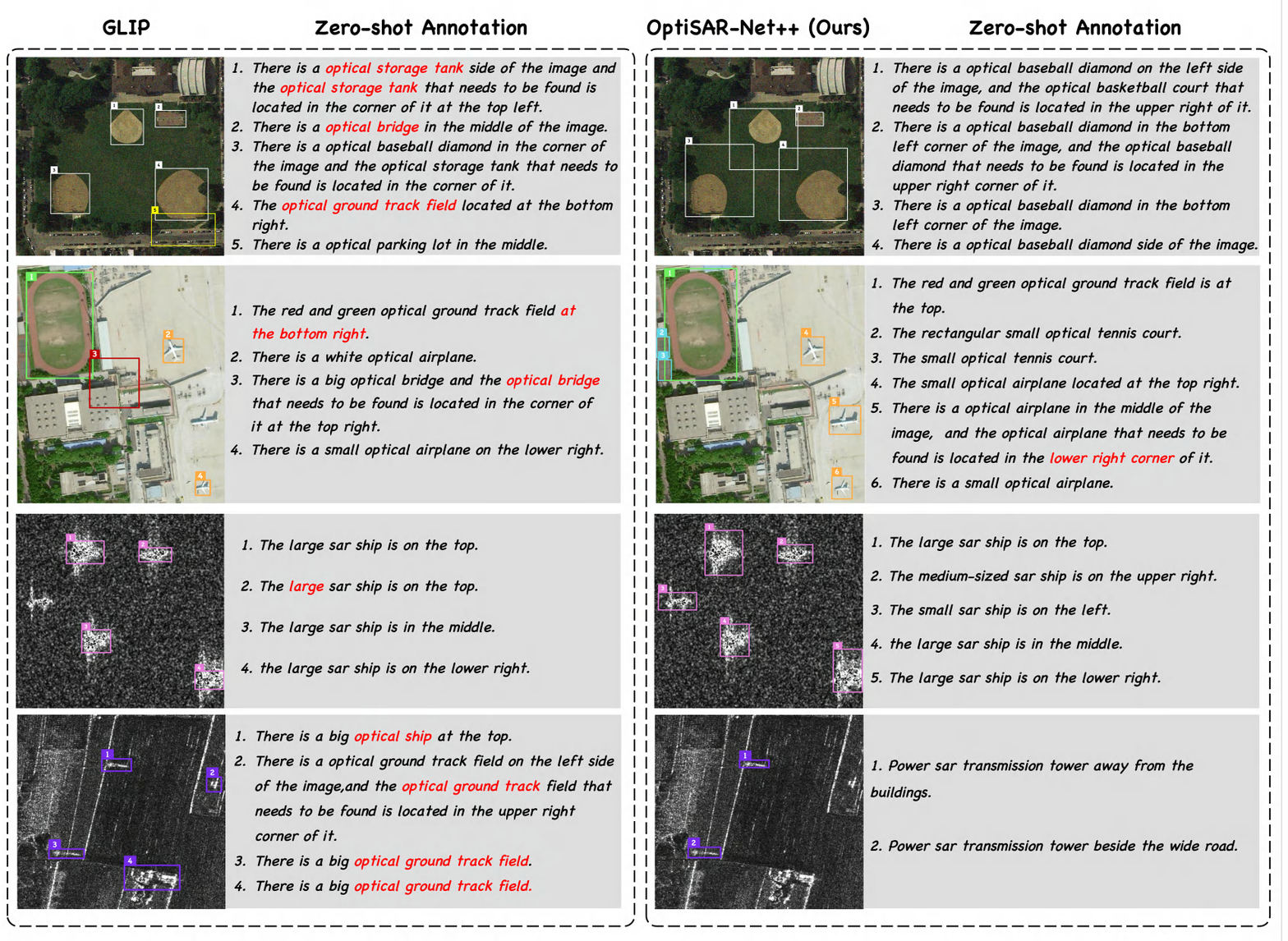}
\caption{Zero-shot image annotation comparison between OptiSAR-Net++ and GLIP. Both methods annotate unseen scenes by matching detections against training text embeddings (confidence $>0.85$). Red text indicates misclassifications. OptiSAR-Net++ outperforms GLIP in category accuracy, attribute discrimination, and spatial-relation richness, demonstrating its potential for automated remote-sensing annotation.}
\label{fig:zeroshot}
\end{figure}

\textit{\textbf{More Visual Grounding Visualizations.}}
Fig.~\ref{fig:moredemo} presents additional predictions in challenging scenes, including dense clusters, severe background clutter, multi-scale objects, and complex spatial queries. Across both optical and SAR domains, OptiSAR-Net++ consistently produces tight, accurate bounding boxes. Particularly in dense scenes, it successfully disambiguates targets among similar candidates using orientation and relational cues, validating the effectiveness of our fine-grained semantic perception mechanism. We also provide several failure cases.

\textit{\textbf{Zero-shot Image Annotation Results.}}
The CLIP-style contrastive paradigm inherently aligns visual and textual spaces, enabling zero-shot transfer. To validate this, we compare OptiSAR-Net++ and GLIP~\cite{glip} on zero-shot image annotation (Fig.~\ref{fig:zeroshot}). We match detected boxes against a unified training-text embedding pool, retaining predictions with confidence $>0.85$.

OptiSAR-Net++ clearly outperforms GLIP in accuracy and completeness. GLIP frequently misclassifies categories (e.g., mistaking a baseball field for an ``oil storage tank'') and generates generic, repetitive descriptions (e.g., labeling all SAR ships as ``large''). Conversely, OptiSAR-Net++ differentiates scale variations (``large'', ``medium'', ``small'') and captures richer orientation cues (``upper left'', ``lower right''). These results demonstrate our framework's strong semantic generalization to novel scenes and categories, offering a promising solution for rapid data annotation and low-resource scene interpretation.

\section{Conclusion and Future Work}
\label{sec:conclusion}

In this paper, we systematically explore cross-domain remote-sensing visual grounding (CD-RSVG). We introduce OptSAR-RSVG, the first large-scale benchmark for this task, and propose OptiSAR-Net++, an efficient Transformer-free framework. By leveraging CLIP-based contrastive learning and adversarial negative mining, OptiSAR-Net++ reframes CD-RSVG from generative regression to efficient retrieval. Furthermore, it integrates PLoRA-MoE for parameter-efficient cross-domain feature decoupling, and TGDF-SSA for precise cross-modal semantic injection and spatial modeling. Extensive experiments demonstrate that OptiSAR-Net++ achieves state-of-the-art performance on the OptSAR-RSVG and DIOR-RSVG benchmarks, showing notable advantages in fine-grained spatial modeling and data efficiency.

However, this work has two main limitations. First, our cross-domain setting is confined to optical and SAR modalities; its generalizability to other data sources (e.g., infrared or multispectral imagery) remains unexplored. Second, compared to Transformer-based models with stronger long-range dependency modeling, our CLIP-based contrastive paradigm may face scalability bottlenecks on larger datasets.

Future work will explore incorporating temporal information from image sequences for dynamic scene understanding (e.g., change detection). Ultimately, we hope the OptSAR-RSVG dataset and OptiSAR-Net++ framework will serve as pivotal foundations for advancing cross-domain remote sensing comprehension and human-machine interaction.

% \section*{Acknowledgments}

% {\appendix[Proof of the Zonklar Equations]
% Use $\backslash${\tt{appendix}} if you have a single appendix:
% Do not use $\backslash${\tt{section}} anymore after $\backslash${\tt{appendix}}, only $\backslash${\tt{section*}}.
% If you have multiple appendixes use $\backslash${\tt{appendices}} then use $\backslash${\tt{section}} to start each appendix.
% You must declare a $\backslash${\tt{section}} before using any $\backslash${\tt{subsection}} or using $\backslash${\tt{label}} ($\backslash${\tt{appendices}} by itself
%  starts a section numbered zero.)}


\bibliographystyle{IEEEtran}
\bibliography{ref}

% \newpage
 
\vspace{11pt}

\begin{IEEEbiographynophoto}{Xiaoyu Tang} (Member, IEEE) received the B.S. degree from South China Normal University, Shanwei, China, in 2003, and the M.S. degree from Sun Yat-sen University, Guangzhou, China, in 2011. 

He is currently pursuing the Ph.D. degree with South China Normal University. He is working with Xingzhi College, South China Normal University, where he is engaged in information system development. His research interests include machine vision, intelligent control, and the Internet of Things. Mr. Tang is a member of the IEEE ICICSP Technical Committee.
\end{IEEEbiographynophoto}

\vspace{-25pt}
\begin{IEEEbiographynophoto}{Jun Dong} (Student Member, IEEE) is currently pursuing the bachelor’s degree in Internet of Things (IoT) engineering with the School of Data Science and Engineering, South China Normal University, Shanwei, China.
\end{IEEEbiographynophoto}

\vspace{-25pt}
\begin{IEEEbiographynophoto}{Jintao Cheng} received his bachelor's degree from the School of Physics and Telecommunications Engineering, South China Normal University, in 2021. He is currently pursuing an MPhil degree at The Hong Kong University of Science and Technology.
\end{IEEEbiographynophoto}

\vspace{-25pt}
\begin{IEEEbiographynophoto}{Rui Fan} (Senior Member, IEEE) received the B.Eng. degree in Automation from the Harbin Institute of Technology in 2015 and the Ph.D. degree in Electrical and Electronic Engineering from the University of Bristol in 2018. He worked as a Research Associate at the Hong Kong University of Science and Technology from 2018 to 2020 and a Postdoctoral Scholar-Employee at the University of California San Diego between 2020 and 2021. He began his faculty career as a Full Research Professor in the College of Electronics \& Information Engineering at Tongji University in 2021. He was promoted to Full Professor in 2022 and attained tenure in 2024, both in the same college and at the Shanghai Research Institute for Intelligent Autonomous Systems. His research interests include computer vision, deep learning, and robotics, with a specific focus on humanoid visual perception under the two-streams hypothesis. Prof. Fan served as an associate editor for ICRA'23/25 and IROS'23/24, an area chair for ICIP'24, and a senior program committee member for AAAI'23/24/25/26. He organized several impactful workshops and special sessions in conjunction with WACV'21, ICIP'21/22/23, ICCV'21/25, and ECCV'22. He was honored by being included in the Stanford University List of Top 2\% Scientists Worldwide between 2022 and 2025, recognized on the Forbes China List of 100 Outstanding Overseas Returnees in 2023, acknowledged as one of Xiaomi Young Talents in 2023, and awarded the Shanghai Science \& Technology 35 Under 35 honor in 2024 as its youngest recipient.
\end{IEEEbiographynophoto}

\vfill

\end{document}